\pdfoutput=1
\documentclass[11pt,table]{article}
\usepackage[utf8]{inputenc}         %
\usepackage[T1]{fontenc}            %
\usepackage{geometry}
\usepackage{libertine}              %
\usepackage[scaled=0.85]{beramono}  %
\usepackage[in]{fullpage}           %
\usepackage[sc]{titlesec}           %
\usepackage{xurl}                   %
\usepackage{amsmath}                %
\usepackage{amsfonts}               %
\usepackage{booktabs}               %
\usepackage[dvipsnames]{xcolor}     %
\usepackage{graphicx}               %
\graphicspath{{figures/}}           %
\usepackage{subcaption}             %
\usepackage{microtype}              %
\usepackage[square]{natbib}         %
\usepackage{siunitx}                %
\usepackage[textsize=scriptsize,textwidth=2.2cm]{todonotes}
\usepackage[para]{footmisc}         %
\makeatletter
\def\blfootnote{\xdef\@thefnmark{}\@footnotetext}
\makeatother

\usepackage{nicefrac}              %
\usepackage{mathtools}             %
\usepackage{amssymb,amsthm}
\usepackage{comment}
\usepackage{etoolbox}
\usepackage{enumitem}              %
\usepackage{tabularx}
\usepackage{longtable}
\usepackage{afterpage}
\usepackage[export]{adjustbox} %
\usepackage{makerobust}
\MakeRobustCommand\rotatebox %
\usepackage{algorithm}
\usepackage[compatible]{algpseudocode} %

\usepackage[pdfusetitle]{hyperref}  %
\usepackage[nameinlink]{cleveref}   %
\newcommand\myshade{60}
\definecolor{mylinkcolor}{HTML}{8F510B} %
\definecolor{mycitecolor}{HTML}{3B4791} %
\definecolor{myurlcolor}{HTML}{8F7F0B}  %
\hypersetup{
    linkcolor  = mylinkcolor!\myshade!black,
    citecolor  = mycitecolor!60!black,
    urlcolor   = myurlcolor!\myshade!black,
    colorlinks = true,
    pdftitle={Training Agents using Upside-Down Reinforcement Learning},
}

\makeatletter
\long\def\@makecaption#1#2{
        \vskip 0.8ex
        \setbox\@tempboxa\hbox{\small {\bf #1:} #2}
        \parindent 1.5em  %
        \dimen0=\hsize
        \advance\dimen0 by -3em
        \ifdim \wd\@tempboxa >\dimen0
                \hbox to \hsize{
                        \parindent 0em
                        \hfil 
                        \parbox{\dimen0}{\def\baselinestretch{0.96}\small
                                {\bf #1.} #2
                                } 
                        \hfil}
        \else \hbox to \hsize{\hfil \box\@tempboxa \hfil}
        \fi
        }
        
\makeatother
\titlespacing{\paragraph}{%
  0pt}{%
  0.5\baselineskip}{%
  1em}%

\def\udrl{Upside-Down Reinforcement Learning}
\def\udRL{Upside-Down RL}
\def\UDRL{UDRL}

\def\xxxn{\rotatebox[origin=c]{180}{RL}}
\def\UDRL{\xxxn{}}

\DeclareMathOperator*{\argmin}{arg\,min}
\DeclareMathOperator*{\len}{len}

\newcommand{\Rab}[2]{\mathbb{R}^{{#1} \times {#2}}}

\title{\vspace{-0.7in}\bf\sc\huge{Training Agents using Upside-Down Reinforcement Learning}}
\author{
    Rupesh Kumar Srivastava$^1$\thanks{Correspondence to: \href{mailto:rupesh@nnaisense.com}{rupesh@nnaisense.com}} \quad 
    Pranav Shyam$^1$\thanks{Now at OpenAI.} \quad 
    Filipe Mutz$^2$\thanks{Now at IFES, Brazil.} \\
    Wojciech Ja\'skowski$^1$ \quad 
    J\"urgen Schmidhuber$^{1,2}$\\
}

\date{{\sc
    $^1$NNAISENSE\\
    $^2$The Swiss AI Lab IDSIA\\
    }
}

\begin{document}
\maketitle

\begin{abstract}
We develop Upside-Down Reinforcement Learning (UDRL or \UDRL{}), a method for learning to act using only supervised learning techniques.
Unlike traditional algorithms, \UDRL{} does not use reward prediction or search for an optimal policy. 
Instead, it trains agents to follow \emph{commands} such as ``obtain so much total reward in so much time.''
Many of its general principles are outlined in a companion report \citep{schmidhuber2019};
the goal of this paper is to develop a practical learning algorithm and show that this conceptually simple perspective on agent training can produce a range of rewarding behaviors for multiple episodic environments.
Experiments show that on some tasks \UDRL{}'s performance can be surprisingly competitive with, and even exceed that of some traditional baseline algorithms developed over decades of research.
Based on these results, we suggest that alternative approaches to expected reward maximization have an important role to play in training useful autonomous agents.
\end{abstract}

\blfootnote{\\\-\hspace{12pt} This version extends work presented at the NeurIPS Deep RL Workshop 2019 \citep{srivastava2019training}.}
\blfootnote{\\\-\hspace{12pt} An demonstrative implementation of our algorithm is available \href{https://colab.research.google.com/drive/1ynS9g7YzFpNSwhva2_RDKYLjyGckCA8H?usp=sharing}{at this link}.}

\section{Introduction}
A recurring theme in Reinforcement Learning (RL) research consists of ideas that attempt to bring the simplicity, robustness and scalability of Supervised Learning (SL) algorithms to traditional RL algorithms.
Perhaps the most popular technique from this class currently is \emph{target networks} \citep{mnih2015} where a copy of an agent's value function is frozen and stored periodically to provide stationary learning targets for temporal-difference learning.
A similar motivation was behind the work of \citet{oh2018}, who used direct SL on past high-return behaviors (imitation learning) to improve exploration in a traditional RL algorithm.
More recently, \citet{arjona2019rudder} used gradient-based SL in recurrent networks to perform contribution analysis in order to map rewards to state-action pairs, thereby making learning from delayed rewards faster.
However, a common thread across most work in this area is that SL subroutines are used to augment traditional RL algorithms, but not replace them completely.

Is it possible to learn to act in high-dimensional environments efficiently using \emph{only} SL, avoiding the issues arising from non-stationary learning objectives common in traditional RL algorithms?
Fundamentally, this appears difficult or even impossible, since feedback from the environment provides \emph{error signals} in SL but \emph{evaluation signals} in RL (see detailed discussions by \citet{rosenstein2004} and \citet{barto2004}).
In other words, an RL agent gets feedback about how useful its actions are, but not about which actions are the best to take in any situation.
On the possibility of turning an RL problem completely into an SL problem, \citet{barto2004} surmised: ``In general, there is no way to do this.''

We develop \emph{\udrl{}} (\UDRL{}) --- a new method of learning that bridges this gap between SL and RL by taking an alternative view on the problem of learning to act.
The goal of learning is no longer to maximize returns in expectation, but to learn to follow \emph{commands} that may take various forms such as ``achieve total reward $R$ in next $T$ time steps'' or ``reach state $S$ in fewer than $T$ time steps''.
We train agents to directly map commands to actions that can fulfill those commands by retrospectively interpreting past experiences as successful examples of following commands.
Such \emph{hindsight}-based learning was initially proposed by \citet{kaelbling1993} and recently revisited by \citet{andrychowicz2017} within the context of traditional RL; here we explore the possibility of relying on it completely.
This leads to two key properties: a) the cumulative reward is an input to the agent rather than a prediction as in value-based RL, and b) learning is based on optimizing a \emph{true} SL objective, unlike many RL algorithms where targets are non-stationary or the data becomes `stale' after a single learning step.
 
The basic principles of \UDRL{} are proposed in a companion technical report by \citet{schmidhuber2019}.
This paper aims to complete this proposal with concrete implementations and discussions in the context of episodic tasks.
Specifically, we show that while \UDRL{} does not formally maximize expected returns, in practice we can train agents that achieve high empirical returns by utilizing the principles of learning to follow commands.
For this purpose we identify minimal additional algorithmic ingredients that guide learning towards commands that correspond to higher returns.

An important consequence of \UDRL{}'s conceptual proximity to SL is that the type of knowledge representation it uses (the \emph{behavior function}; see next section) should in principle be able to exploit the learning and generalization capabilities of powerful function approximators such as neural networks without additional stabilization techniques such as \emph{target networks}.
We demonstrate that this knowledge representation can indeed be realized in practice in several non-trivial environments.
Finally, we show two interesting properties of \UDRL{} agents: they can naturally learn from delayed rewards, and they can adjust their behavior \emph{after training} in response to commands specifying varying amounts of desired returns.

\section{\udrl{}}

\paragraph{Terminology \& Notation}
In what follows, $s$, $a$ and $r$ denote \emph{state}, \emph{action}, and \emph{reward} respectively.
We assume $s \in \mathcal{S}$ and $a \in \mathcal{A}$ where $\mathcal{S}$ and $\mathcal{A}$ are finite sets whose sizes depend on the environment, and consider episodic Markovian environments with scalar rewards ($r \in \mathbb{R}$).
This simplifies exposition,\footnote{We can use probabilities and states, instead of probability densities and observations.} but the general principles of \UDRL{} are not limited to these settings.
A \emph{policy} $\pi: \mathcal{S} \rightarrow \mathcal{A}$ is a function that selects an action in a given state.
A stochastic policy maps a state to a probability distribution over actions.
Each \emph{episode} consists of an agent's interaction with its environment starting in an initial state and ending in a terminal state while following any policy.
Right subscripts denote time indices (e.g. $s_t, t \in \mathbb{N}^0$). 
A \emph{trajectory} $\tau$ is the sequence $\langle(s_t, a_t, r_t, s_{t+1})\rangle,  t = 0, \dots, T-1$ describing an episode of length $T$.
A subsequence of a trajectory is a \emph{segment} (denoted $\kappa$), and
\emph{return} refers to the cumulative reward over a segment.

\begin{figure*}[t]
    \centering
    \includegraphics[width=0.7\textwidth]{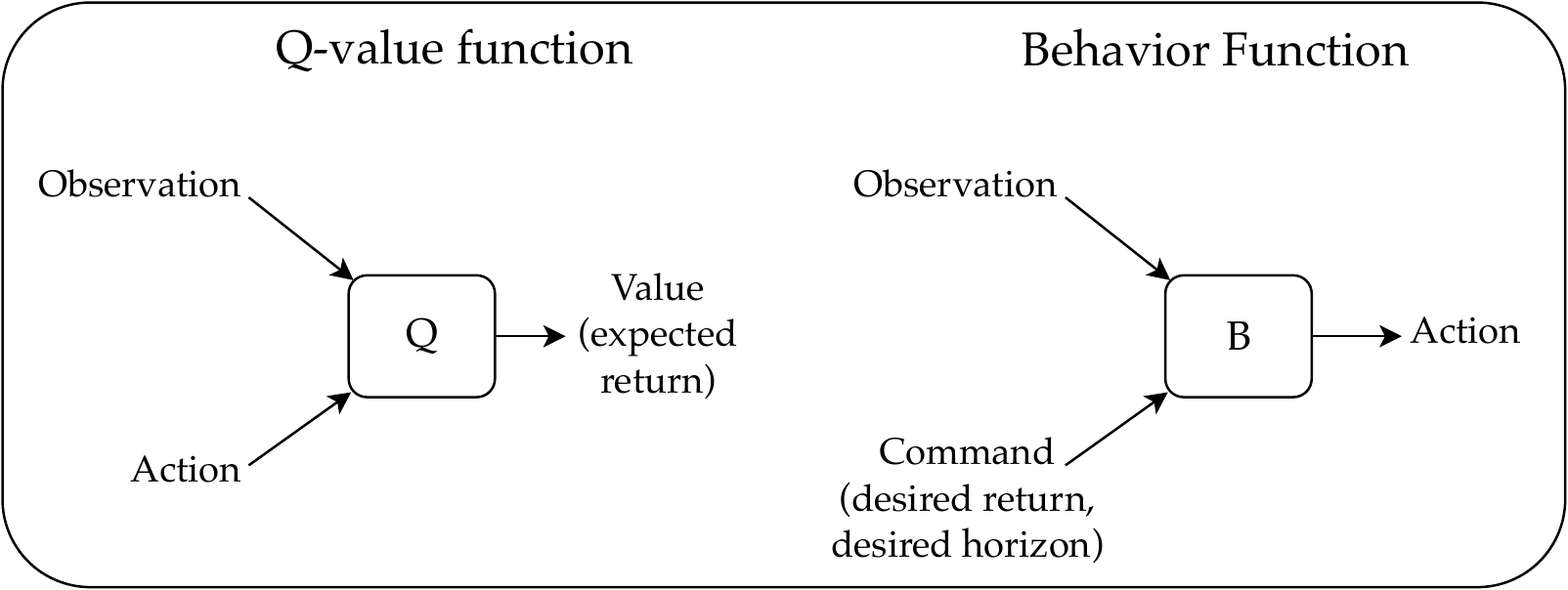}
    \caption{The action-value function ($Q$) in traditional RL predicts expected return over the long-term future while the behavior function ($B$) in \UDRL{} takes commands such as desired future return as input --- conceptually the roles of actions and returns are switched. $B$ directly produces a probability distribution over immediate actions. In addition, it may have other command inputs such as desired states or a desired time horizon.}
    \label{fig:comparison}
\end{figure*}

\subsection{Knowledge Representation}
\label{sec:basics}

In many traditional model-free RL algorithms, the core component of a learning agent is a \emph{value function} that encodes knowledge about how the environment will reward actions taken according to its current policy.
Another common class of algorithms uses direct policy search to optimize returns, in which case the policy represents all of the agent's knowledge.
Currently popular RL algorithms typically combine both of these approaches by using a learned value function to improve policy search.

In contrast, a basic \UDRL{} agent neither predicts values nor explicitly maximizes them, but instead only learns to interpret \emph{commands} given at each time step to determine the appropriate action to be taken.
Commands can specify requirements for desired behavior such as total return, desired goals to reach, or additional constraints.
Throughout this paper, we only consider commands of the type: ``achieve a desired return $d^r$ in the next $d^h$ (desired horizon) steps from the current state''.
Given a particular definition of commands, a \emph{behavior function} is trained using SL to encode knowledge about all of the agent's past interactions with the environment.
It can then be used to produce actions in response to a variety of commands during deployment or for exploration.
\autoref{fig:comparison} schematically illustrates that conceptually, the behavior function is an upside-down version of the action-value function commonly used in conventional RL algorithms, since the roles of actions and returns are switched.
Additionally, note that value functions predict the expected value of returns, while the input to a behavior function is not an expected value.

Formally, let the agent be given a command indicating that a return of $d^r$ is desired in the next $d^h$ steps.
Let $S$ and $A$ be random variables denoting the environment's state and the agent's next action respectively.
Let $R_{d^h}$ be the random variable denoting the return obtained by the agent during the next $d^h$ time steps.
Then the behavior function $B_\pi$ produces the conditional probability that taking an action $a \in \mathcal{A}$ and thereafter acting using the policy $\pi$ leads to the desired return in the desired horizon:

\begin{align}
B_\pi(a, s, d^r, d^h) &= P(A = a \mid R_{d^h} = d^r, S = s; \pi)\label{eq:bf-formal1}\\
                      &= \dfrac{P(A = a, R_{d^h} = d^r, S = s; \pi)}{P(R_{d^h} = d^r, S = s; \pi)}.\label{eq:bf-formal2}
\end{align}

Assume that instead of a policy, only a dataset of trajectories $\mathcal{T}$ generated using an unknown policy is available, and we are interested in the behavior function $B_\mathcal{T}$ for the dataset.
With discrete states and actions, it can be estimated from the available data (following \autoref{eq:bf-formal2}) as

\begin{equation}\label{eq:bf-simple}
B_\mathcal{T}(a, s, d^r, d^h) = \dfrac{N^{a}_\kappa(s, d^r, d^h)}{N_\kappa(s, d^r, d^h)},
\end{equation}
where $N_\kappa(s, d^r, d^h)$ is the number of trajectory segments in $\mathcal{T}$ that start in state $s$, have length $d^h$ and total reward $d^r$, and $N^{a}_\kappa(s, d^r, d^h)$ is the number of such segments where the first action was $a$.

\begin{figure}[t]
\centering
\begin{minipage}[b]{0.495\textwidth}
    \centering
    \includegraphics[width=0.65\textwidth]{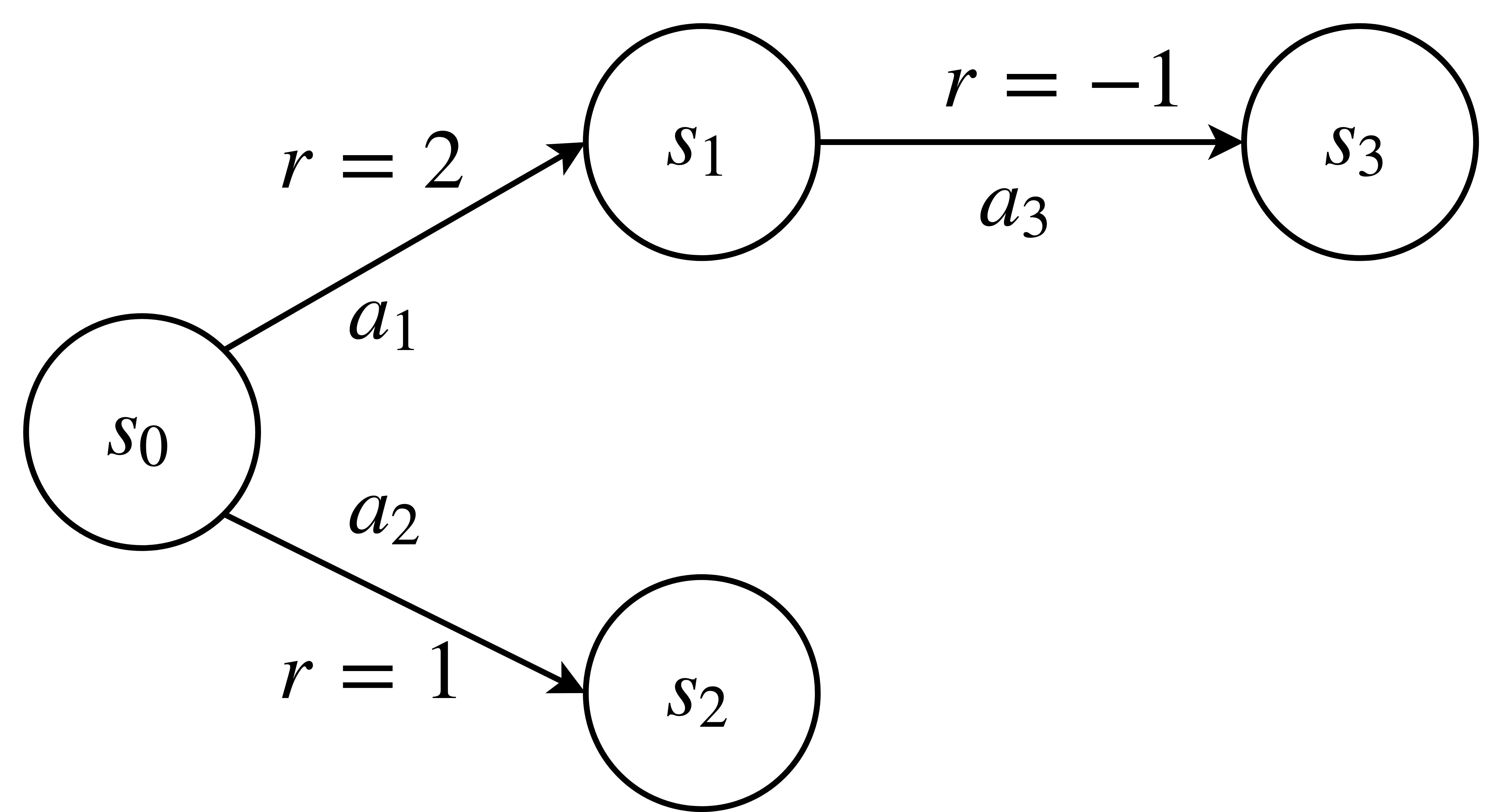}
    \caption{A toy environment with four states.}\label{fig:toyenv}
\end{minipage}
\begin{minipage}[b]{0.495\textwidth}
    \centering\small
    \captionsetup{type=table} %
    \caption{\noindent A behavior function based on all unique trajectories for the toy environment (left image).}\label{tab:toybt}
    \begin{tabular}[b]{@{} c S[table-format=1] S[table-format=1] | c @{}}
    \toprule
    State  & {Desired Return} & {Desired Horizon} & Action \\
    \midrule
    $s_0$  &  2 & 1 & $a_1$   \\
    $s_0$  &  1 & 1 & $a_2$   \\
    $s_0$  &  1 & 2 & $a_1$   \\ 
    $s_1$  & -1 & 1 & $a_3$  \\
    \bottomrule
\end{tabular}
\end{minipage}
\end{figure}

Consider the deterministic toy environment in \autoref{fig:toyenv} in which all trajectories start in $s_0$ or $s_1$ and end in $s_2$ or $s_3$.
When $\mathcal{T}$ is the set of all unique trajectories in this environment (there are just three), $B_\mathcal{T}$ can be expressed in a simple tabular form in \autoref{tab:toybt}.
Intuitively, it answers the question: ``if an agent is in a given state and desires a given return over a given horizon, which action should it take next based on past experience?''.
In this toy example, it was sufficient to specify the only action that achieves each command, but it should be noted that in general, $B_\mathcal{T}$ must produce a probability distribution over actions even in deterministic environments, since there may be multiple immediate actions compatible with the same command and state.
For example, this would be the case in the toy environment if the transition $s_0 \rightarrow s_2$ had a reward of 2.

Given a set of trajectories from an environment (these may be produced using random actions at the beginning of learning), an agent can use $B_\mathcal{T}$ to select actions when given external commands.
However, it immediately becomes clear that using \autoref{eq:bf-simple} is both computationally expensive and conceptually limited.
Computing the behavior function for every command involves a rather expensive search procedure over the entire dataset.
Moreover, with limited experience or in continuous-valued state spaces, it is likely that no examples of a segment with the queried $s, d^r$ and $d^h$ exist, even though intuitively there is significant problem-specific structure in the agent's experience that can be exploited to generalize in such situations.
For example, after hitting a ball a few times and observing its resulting velocity, an agent should be able to understand that in order to achieve a higher velocity, it should hit the ball with a force greater than before.

The solution is to \emph{learn} a function to approximate $B_\mathcal{T}$ that compresses the agent's experience, makes computation of the conditional probabilities efficient and enables generalization to unseen states or commands.
Using a loss function $L$, it can be estimated by solving the following SL problem:

\begin{align}\label{eq:bf}
B_{\mathcal{T}} =& \argmin_B \sum_{(\tau,\,t_1,\,t_2)} L(B(a_{t_1}, s_{t_1}, d^r, d^h), a_{t_1}),\\
\text{where }& 0 \leq t_1 < t_2 \leq \len(\tau) \text{ for } \tau \in \mathcal{T}, d^r = \sum_{t=t_1}^{t_2} r_t \text{ and } d^h = t_2 - t_1.\nonumber
\end{align}

Here $\len(\tau)$ is the length of any trajectory $\tau$, and in general $-L$ can be any proper score function \citep{gneiting2007strictly}.
In practice, we assume an output distribution over actions based on the environment,\footnote{$B$ produces a valid probability for each action in the discrete action setting (\autoref{eq:bf-simple}). For continuous-valued actions, $B$ should produce a probability density instead, and the action distribution becomes a modeling choice. We chose independent Normal distributions in this case to be consistent with the traditional RL algorithms used for comparisons.}
parameterize $B$ accordingly, and use the cross-entropy between the observed and predicted action distributions as the loss function.
Sampling input-target pairs for training is simple: sample a segment bounded by time indices $t_1$ and $t_2$ from any trajectory, then construct training data by taking its first state $(s_{t_1})$ and action $(a_{t_1})$, and compute the values of $d^r$ and $d^h$ for it in hindsight \citep{kaelbling1993,andrychowicz2017,rauber2017}.

\subsection{Maximizing Episodic Returns using Upside-Down RL}

We now focus on a common problem setting where we wish to train an agent to achieve the maximum possible average episodic returns.
Using a concrete algorithm applied to benchmarks in this setting, our goal is to understand whether useful behavior functions can be learned in practice, and whether a simple off-policy algorithm based purely on SL can solve interesting RL problems.

High-level pseudo-code for the proposed algorithm is described in \autoref{algo:udrl-high}.
It starts by initializing an empty replay buffer to collect the agent's experiences during training, and filling it with a few episodes of random interactions.
The behavior function of the agent is continually improved by supervised training on previous experiences recorded in the replay buffer.
After each training phase, the behavior function is used to act in the environment to obtain new experiences that are added to the replay buffer.
This procedure continues until a stopping criterion is met, such as reaching the allowed maximum number of interactions with the environment.
The remainder of this section describes each step of the algorithm and introduces the hyperparameters.
A concise list of hyperparameters is also provided in \autoref{sec:hypers}.

\begin{algorithm}[t]
	\caption{\udrl{}: High-level Description.}
	\label{algo:udrl-high}
	\begin{algorithmic}[1]
 		\STATE Initialize replay buffer with warm-up episodes using random actions \COMMENT{\autoref{subsec:replay}}
 		\STATE Initialize a behavior function \COMMENT{\autoref{subsec:behaviorfunc}}
		\WHILE{stopping criteria is not reached}
    		\STATE Improve the behavior function by training on replay buffer \COMMENT{\autoref{subsec:training}}
            \STATE Sample exploratory initial commands based on replay buffer\COMMENT{\autoref{subsec:cmd_proposal}}
            \STATE Explore using sampled commands in \autoref{algo:behavior} and add to replay buffer \COMMENT{\autoref{subsec:act}}
            \STATE If evaluation is required, evaluate using evaluation commands in \autoref{algo:behavior}  \COMMENT{\autoref{subsec:eval}}
        \ENDWHILE
	\end{algorithmic}
\end{algorithm}

\subsubsection{Replay Buffer}
\label{subsec:replay}

The set of trajectories $\mathcal{T}$ used for training the behavior function are stored in a replay buffer.
\UDRL{} does not explicitly maximize returns, but learning can be biased towards  higher returns by selecting the trajectories on which the behavior function is trained.
To do so we use a replay buffer with the best $Z$ trajectories seen so far, where $Z$ is a fixed hyperparameter.
The trade-off is that at the end of training, the agent may not reliably obey commands desiring low returns, which the behavior function only sees at the beginning of training.
At the beginning of training, an initial set of trajectories is generated by executing random actions in the environment (\autoref{algo:udrl-high}; Line 1).

\subsubsection{Behavior Function}
\label{subsec:behaviorfunc}

\begin{algorithm}[t]
	\caption{Generates an Episode using the Behavior Function.}
	\label{algo:behavior}
	\begin{algorithmic}[1]
        \REQUIRE{Initial command $c_0 = (d_0^r, d_0^h)$, Initial state $s_0$}, Behavior function $B$ parameterized by $\theta$
        \ENSURE{Episode data $E$}
        \STATE $E \leftarrow \varnothing$
        \STATE $t \leftarrow 0$
        \WHILE{episode is not done}
            \STATE Compute $P(a_t \mid s_t, c_t) = B(s_t, c_t; \theta)$
            \STATE Execute $a_t \sim P(a_t \mid s_t, c_t)$ to obtain reward $r_t$ and next state $s_{t+1}$ from the environment
            \STATE Append $(s_t, a_t, r_t)$ to $E$
            \STATE $t \leftarrow t + 1$
            \STATE $d_t^r \leftarrow d_{t-1}^r - r_{t-1}$ \COMMENT{Update desired return\hphantom{h}}
            \STATE $d_t^h \leftarrow d_{t-1}^h - 1$ \COMMENT{Update desired horizon}
            \STATE $c_t \leftarrow (d_t^r, d_t^h)$
        \ENDWHILE
	\end{algorithmic}
\end{algorithm}

At any time $t$ during an episode, the current behavior function $B$ produces a distribution over actions $P(a_t \mid s_t, c_t) = B(s_t, c_t; \theta)$ for the current state $s_t$ and command $c_t \coloneqq (d_t^r, d_t^h)$, where $d_t^r \in \mathbb{R}$ is the \emph{desired return}, $d_t^h \in \mathbb{N}$ is the \emph{desired horizon}, and $\theta$ is a vector of trainable parameters initialized randomly at the beginning of training.
Neural networks are a natural choice for representing the behavior function due to their high capacity and flexibility, but other function approximators may be a better choice in some contexts.
Given an \textbf{initial command} $c_0$ for a new episode, a new trajectory is generated using \autoref{algo:behavior} by sampling actions according to $B$ and \textbf{updating the current command using the obtained rewards and time left} at each time step until the episode terminates.
It is notable that $B$ is not used to act like a typical policy, instead it induces a policy at each step.

We note two implementation details that only affect agent evaluations (not training): $d_t^h$ is always set to $\max(d_t^h, 1)$ such that it is a valid time horizon, and $d_t^r$ is clipped such that it is upper-bounded by (an estimate of) the maximum return achievable in the environment.
Clipping the desired return during evaluation avoids situations where negative rewards ($r_t$) can lead to desired returns that are not achievable from any state (see \autoref{algo:behavior}; line 8).
\emph{LunarLander} is the only common RL benchmark where this behavior was observed.

\subsubsection{Training the Behavior Function}
\label{subsec:training}

As discussed in \autoref{sec:basics}, $B$ is trained using supervised learning on input-target examples from any past episode by minimizing the loss in \autoref{eq:bf}.
The goal of training is to make the behavior function produce outputs consistent with all trajectories in the replay buffer.
To draw a training example from a random episode in the replay buffer, time step indices $t_1$ and $t_2$ are selected randomly such that $0 \leq t_1 < t_2 \leq T$, where $T$ is the length of the selected episode.
Then the input for training $B$ is $(s_{t_1}, (d^r, d^h))$, where $d^r = \sum_{t=t_1}^{t_2} r_t$ and $d^h = t_2 - t_1$, and the target is $a_{t_1}$, the action taken at $t_1$.

Several heuristics may be used to select and combine training examples into mini-batches for gradient-based SL.
For all experiments in this paper, \textbf{only trailing segments were sampled from each episode}, i.e.,\ we set $t_2 = T-1$ where $T$ is the length of any episode.
This discards a large amount of potential training examples but is a good fit for episodic tasks where the goal is to optimize the total reward until the end of each episode.
It also makes training easier, since the behavior function only needs to learn to execute a subset of possible commands.
To keep the setup simple, a fixed number of training iterations using Adam \citep{kingma2014} were performed in each training step for all experiments.

\subsubsection{Sampling Exploratory Commands}
\label{subsec:cmd_proposal}

After each training phase the agent can be given new commands, potentially achieving higher returns due to additional knowledge gained by further training.
To profit from such \emph{exploration through generalization}, a set of new initial commands $c_0$ to be used in \autoref{algo:behavior} is generated.
\begin{enumerate}
    \item A number of episodes with the highest returns are selected from the replay buffer. This number is a hyperparameter and remains fixed during training.
    \item The exploratory desired horizon $d^h_0$ is set to the mean of the lengths of the selected episodes.
    \item The exploratory desired returns $d^r_0$ are sampled from the uniform distribution $\mathcal{U}[M, M+S]$ where $M$ is the mean and $S$ is the standard deviation of the selected episodic returns. 
\end{enumerate}
This procedure was chosen due to its simplicity and ability to adjust the agent's optimism using a single hyperparameter.
Intuitively, it tries to generate episodes (aided by stochasticity) with returns similar or better than the highest return episodes in the replay buffer.
For higher dimensional commands, such as those specifying target states, different strategies that follow similar ideas can be designed.
\citet{schmidhuber2019} postulated that any variety of heuristics may be used here, but in practice it is very important to select exploratory commands that lead to data that is meaningfully different from existing experience to drive learning progress.
An inappropriate exploration strategy can lead to very slow or stalled learning.

\subsubsection{Generating Experience}
\label{subsec:act}

Once the exploratory commands are sampled, it is straightforward to generate new exploratory episodes of interaction by using \autoref{algo:behavior}, which works by repeatedly sampling from the action distribution predicted by the behavior function and updating its inputs for the next step.
A fixed number of episodes are generated in each iteration of learning, and added to the replay buffer.

\subsubsection{Evaluation}
\label{subsec:eval}

\autoref{algo:behavior} is also used to evaluate the agent at any time using evaluation commands derived from the most recent exploratory commands.
The initial desired return $d^r_0$ is set to $M$, the lower bound of the desired returns from the most recent exploratory command, and the initial desired horizon $d^h_0$ is reused.
This strategy assumes that returns/horizons similar to the observed commands are repeatable with high probability in the environment, which is valid for many environments of practical interest.
However, it does not apply to arbitrarily stochastic environments where it may be necessary to explicitly model the relationship between states and ``valid'' commands; we leave this to future work.
When evaluating using continuous-valued actions, we follow convention and use the mode of the action distribution.

\section{Experiments}
\label{sec:experiments}

Our experiments are designed to a) determine the practical feasibility of \UDRL{}, and, b) put its performance in context of traditional RL algorithms.
We compare to Double Deep Q-Networks (DQN~\cite{mnih2015,van2016deep}) and Advantage Actor-Critic (A2C; synchronous version of the algorithm proposed by \citet{mnih2016}) for environments with discrete actions, and TRPO \citep{schulman2015trust}, PPO \citep{schulman2017proximal} and DDPG \citep{lillicrap2015continuous} for environments with continuous actions.
These algorithms are recent precursors of the current state-of-the-art, embody the principles of value prediction and policy gradients from which \UDRL{} departs, and derive from a significant body of research.

All agents were implemented using fully-connected feed-forward neural networks, except for TakeCover-v0 where we used convolutional networks.
The command inputs were multiplied by a scaling factor kept fixed during training.
Simply concatenating commands with states lead to very inconsistent results, making it extremely difficult to find good hyperparameters. 
We found that use of architectures with \emph{fast weights} \citep{schmidhuber1992learning} --- where outputs of some units are weights of other connections --- considerably improved reliability over several runs.
Such networks have been used for RL in the past \citep{gomez2005evolving}, and can take a variety of forms.\footnote{See \autoref{sec:app-netarchs} for further discussion on fast weights.}
We included two of the simplest choices in our hyperparameter search: \emph{gating} (as commonly used in LSTM \citep{hochreiter1997long}) and \emph{bilinear} \citep{jayakumar2020multiplicative}, using them only at the first layer in fully-connected networks and at the last layer in convolutional ones. This design makes it difficult to ignore command inputs during training.

We use environments with both low and high-dimensional (visual) observations, and both discrete and continuous-valued actions: LunarLander-v2 based on Box2D~\citep{catto}, TakeCover-v0 based on VizDoom~\citep{kempka2016} and Swimmer-v2 \& InvertedDoublePendulum-v2 based on the MuJoCo~\citep{todorov2012mujoco} simulator, available in the Gym library~\citep{brockman2016}.
These environments are useful for benchmarking but represent solved problems so our goal is not to obtain the best possible performance, but to ensure rigorous comparisons.
The challenges of doing so for deep RL experiments have been highlighted recently by \citet{henderson2018deep} and \citet{colas2018many}.
We follow their recommendations by using separate seeds for training and evaluation and 20 independent experiment runs for all final comparisons.
We also used a hyperparameter search to tune each algorithm on each environment (see supplementary materials for details).
Agents were trained for 10\,M environmental steps for LunarLander-v2/TakeCover-v0 and evaluated using 100 episodes at 50 K step intervals. 
For the less stochastic Swimmer-v2 and InvertedDoublePendulum-v2, training used 5\,M steps and evaluation used 50 episodes to reduce the computational burden.
The supplementary material includes further details of environments, architectures, hyperparameter tuning and the benchmarking setup.
A notebook fully reproducing the training on LunarLander-v2 is available \href{https://colab.research.google.com/drive/1ynS9g7YzFpNSwhva2_RDKYLjyGckCA8H?usp=sharing}{at this link}.

\begin{figure*}[t]
    \centering
    \begin{subfigure}[t]{0.45\textwidth}
    \includegraphics[width=\textwidth]{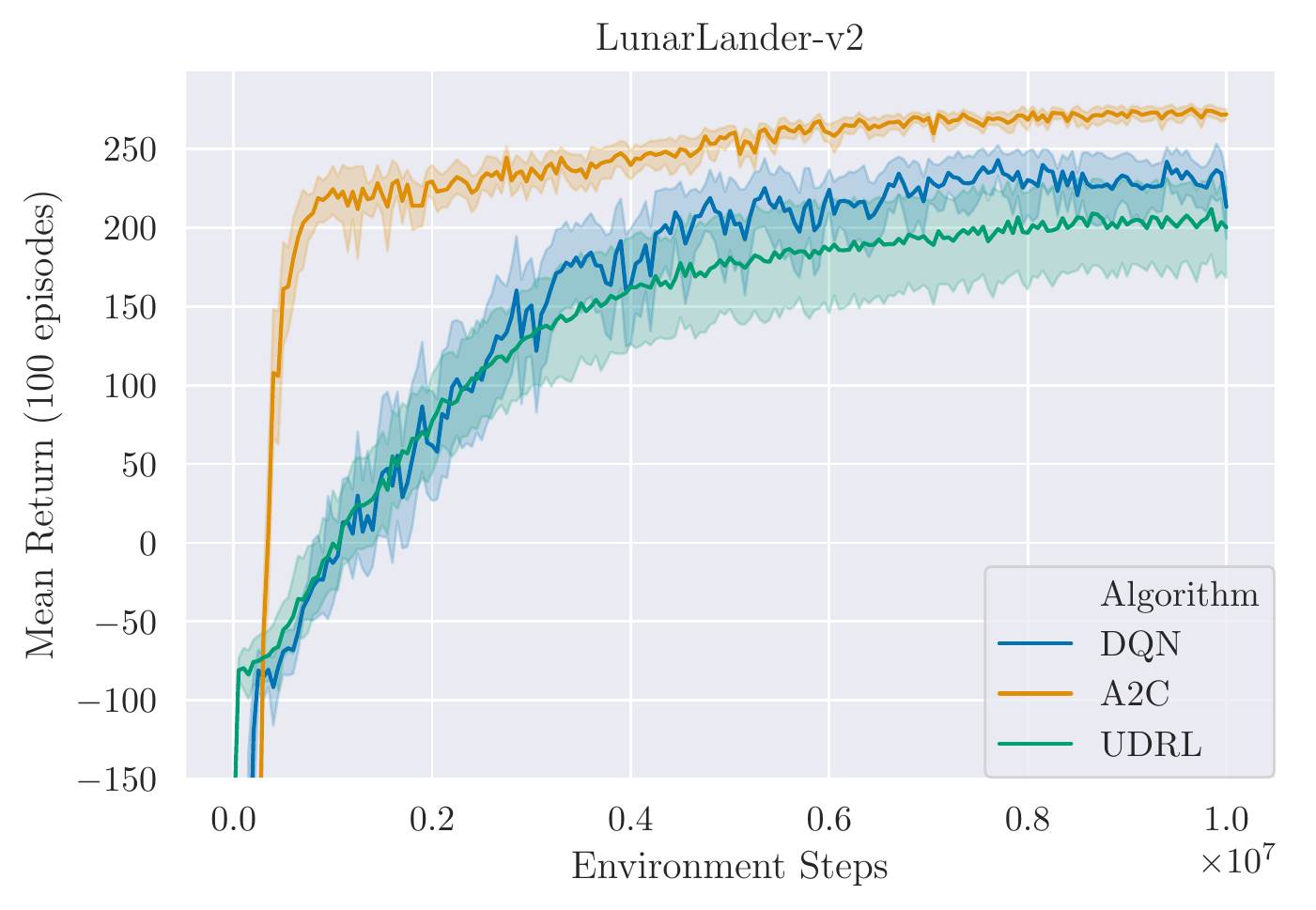}
    \end{subfigure}
    ~
    \begin{subfigure}[t]{0.45\textwidth}
    \includegraphics[width=\textwidth]{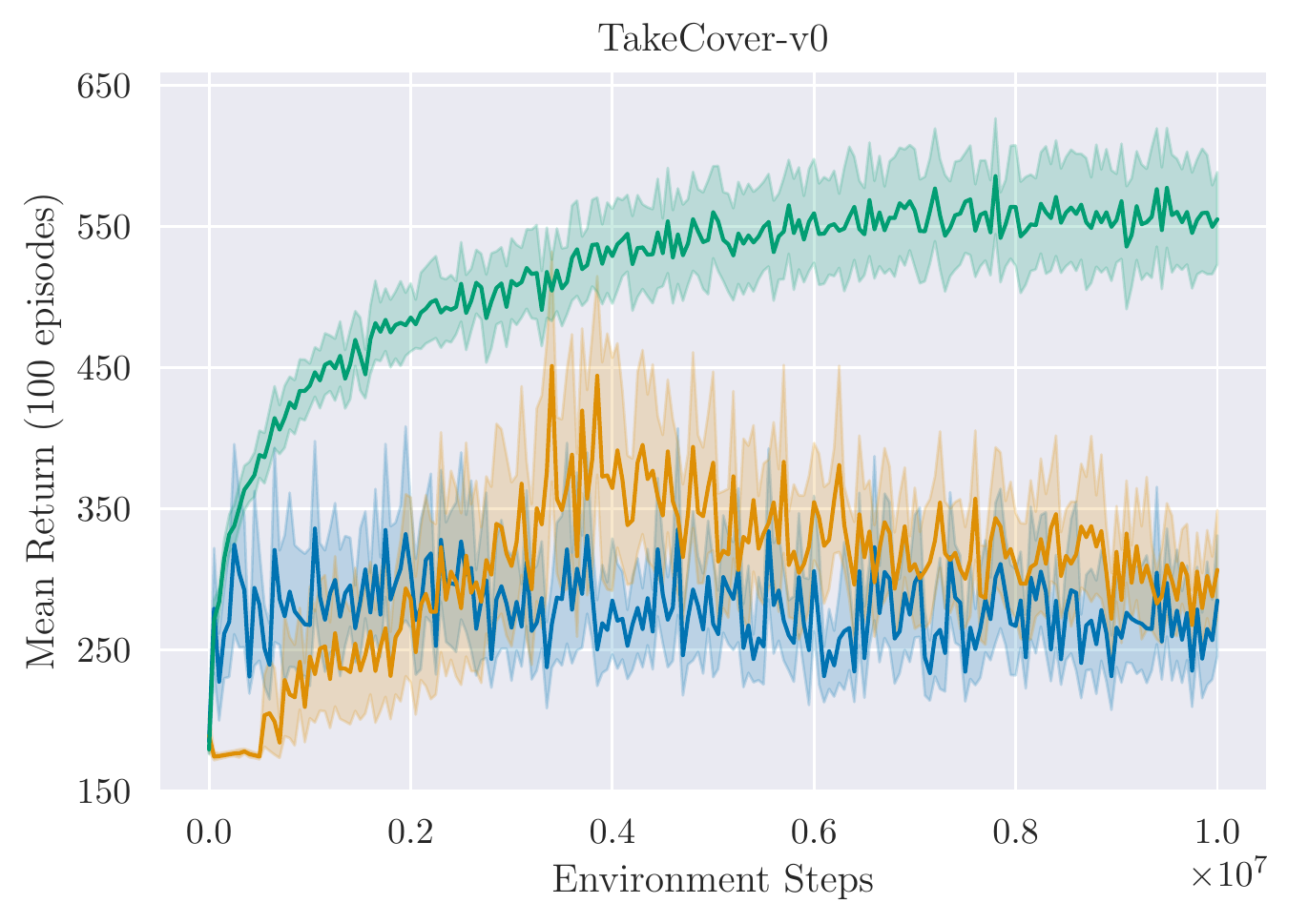}
    \end{subfigure}
    \\
    \begin{subfigure}[t]{0.45\textwidth}
    \includegraphics[width=\textwidth]{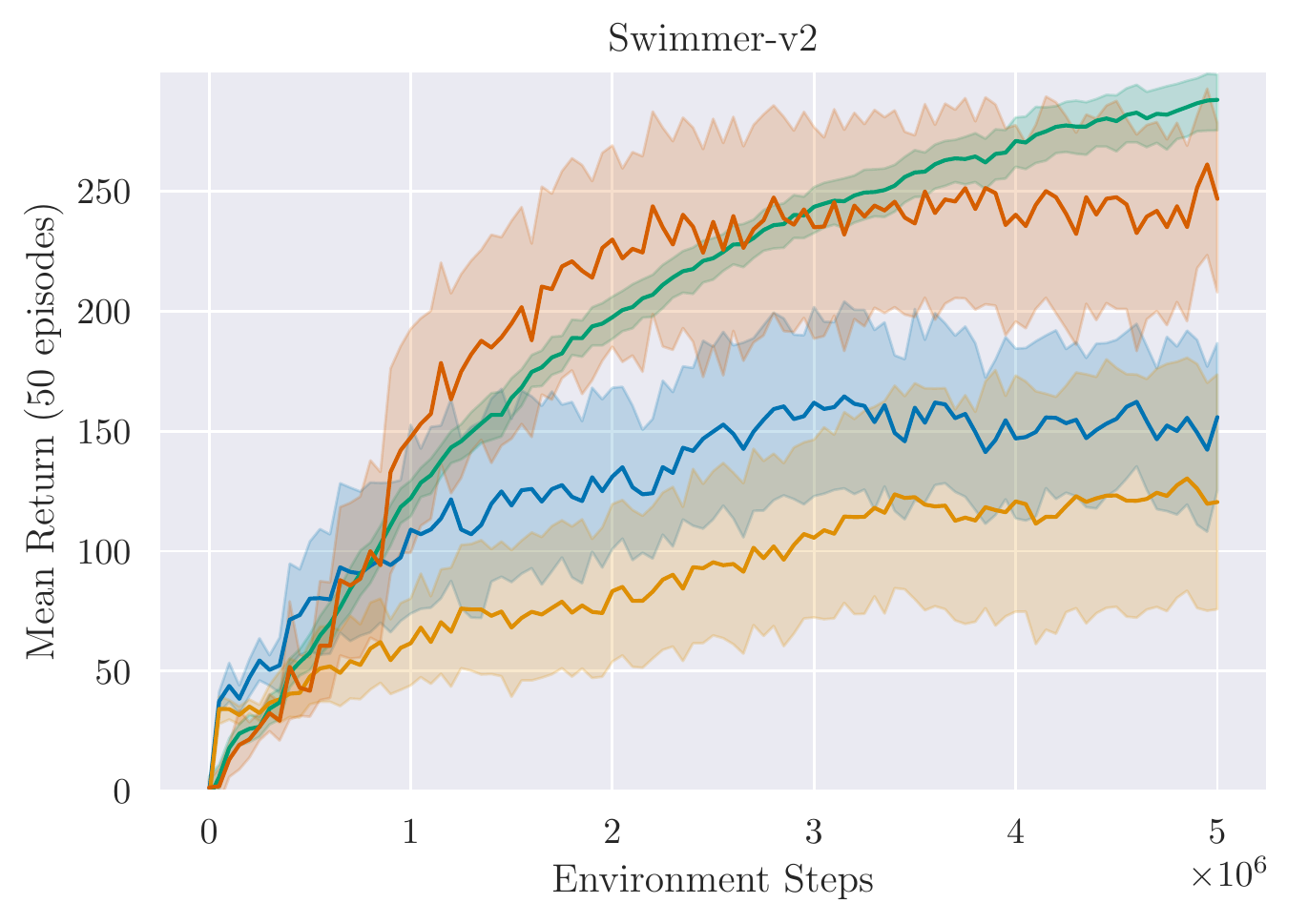}
    \end{subfigure}
    ~
    \begin{subfigure}[t]{0.45\textwidth}
    \includegraphics[width=\textwidth]{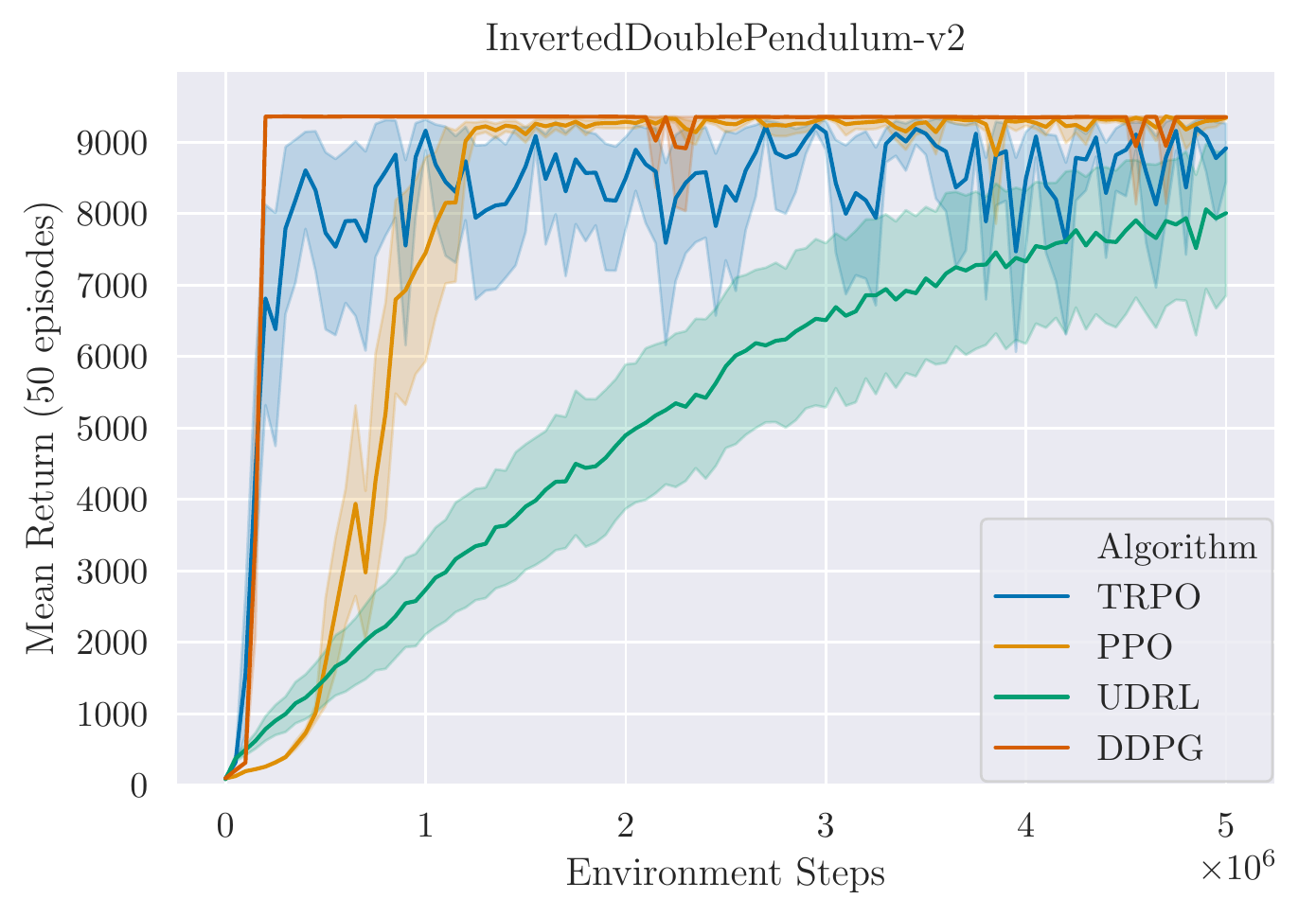}
    \end{subfigure}

    \caption{Results on tasks with discrete-valued actions (top row) and continuous-valued actions (bottom row). Solid lines represent the mean of evaluation scores over 20 runs using tuned hyperparameters and experiment seeds 1--20. Shaded regions represent 95\% confidence intervals using 1000 bootstrap samples. \UDRL{} is competitive with or outperforms traditional baseline algorithms on all tasks except InvertedDoublePendulum-v2.}
    \label{fig:ll_and_tc}
\end{figure*}

\subsection{Results}
\label{subsec:results}

The final 20 runs are plotted in aggregate in \autoref{fig:ll_and_tc}, with dark lines showing the mean evaluation return and shaded regions indicating 95\% confidence intervals with 1000 bootstrap samples.
On LunarLander-v2, all algorithms successfully solved the task (reaching average returns over 200). 
\UDRL{} performed similar to DQN but both algorithms were behind A2C in sample complexity and final returns, which benefits from its use of multi-step returns.
On TakeCover-v0, \UDRL{} outperformed both A2C and DQN comfortably.
Inspection of the individual evaluation curves showed that both baselines had highly fluctuating evaluations, sometimes reaching high scores but immediately dropping in performance at the next evaluation.
This high variability led to overall lower scores when aggregated.
While it may be possible to address these instabilities by incorporating additional techniques or modifications to the environment, it is notable that our simple implementation of \UDRL{} does not suffer from them.

On the Swimmer-v2 benchmark, \UDRL{} outperformed TRPO and PPO, and was on par with DDPG.
However, DDPG's evaluation scores were highly erratic (indicated by large confidence intervals), and it was rather sensitive to hyperparameter choices.
It also stalled completely at low returns for a few random seeds, while \UDRL{} showed consistent progress.
Finally, on InvertedDoublePendulum-v2, \UDRL{} was much slower in reaching the maximum return on average, compared to other algorithms which typically solved the task within 1\,M steps.
While most runs did reach the maximum return ($\approx$9300), some failed to solve the task within the step limit and one run stalled at the beginning of training. 

Overall, the results show that while \UDRL{} can currently lag behind traditional RL algorithms on some tasks, it can also outperform them on other tasks, despite its simplicity and relative immaturity. 
The next section shows that it can be even more effective when the reward function is particularly challenging.

\begin{figure*}[t]
    \centering
    \begin{subfigure}[b]{0.32\textwidth}
    \includegraphics[width=\textwidth]{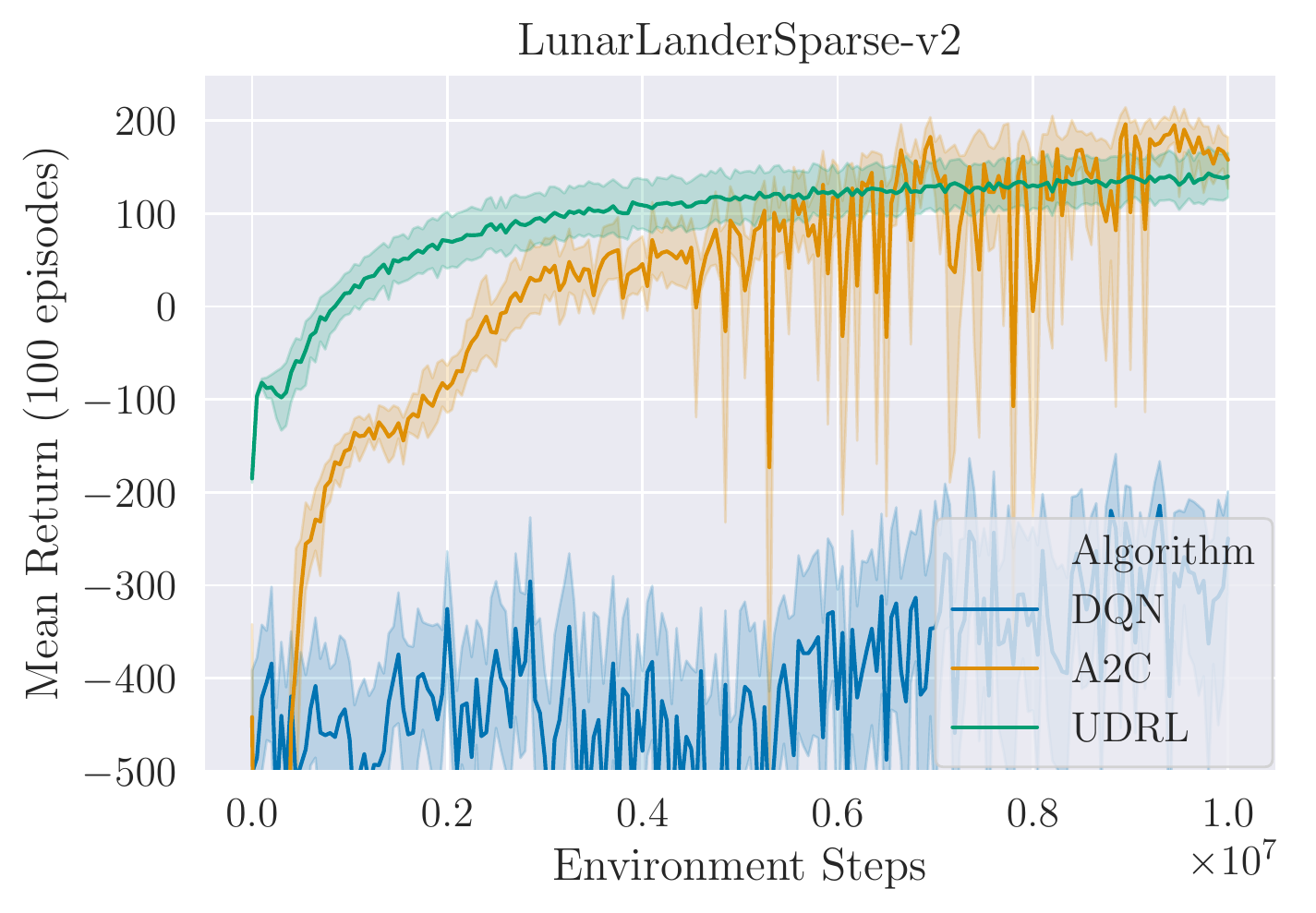}
    \end{subfigure}
    ~
    \begin{subfigure}[b]{0.32\textwidth}
    \includegraphics[width=\textwidth]{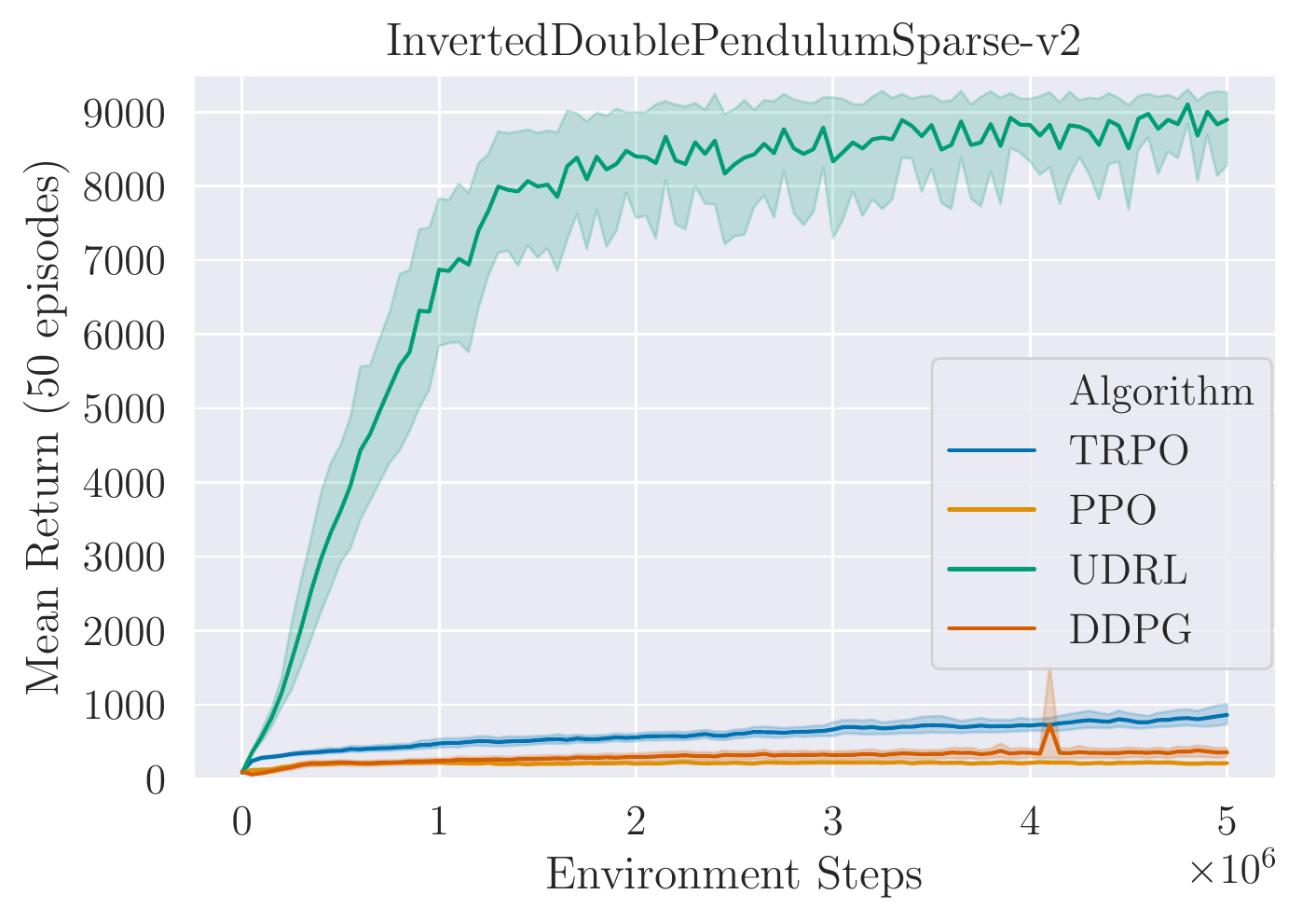}
    \end{subfigure}
    ~
    \begin{subfigure}[b]{0.32\textwidth}
    \includegraphics[width=\textwidth]{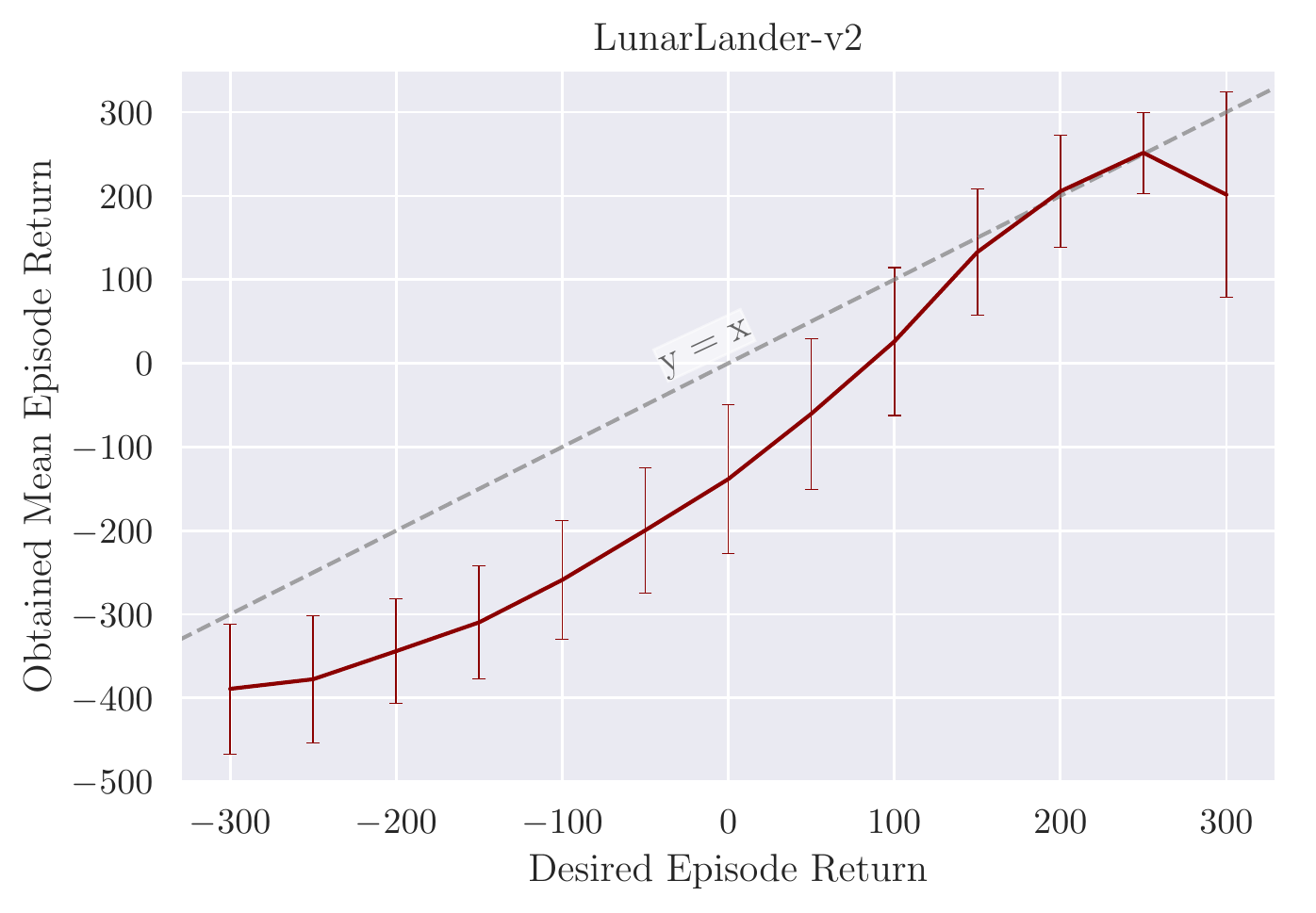}
    \end{subfigure}
    \caption{{\bf Left/Middle}: Results on sparse delayed reward versions of benchmark tasks, with same semantics as \autoref{fig:ll_and_tc}. 
    A2C with 20-step returns was the only baseline to reach good performance on LunarLanderSparse-v2 (see main text).
    SwimmerSparse-v2 results are included in the appendix.
    {\bf Right:} Desired vs.\ obtained returns from a trained \UDRL{} agent, showing ability to adjust behavior in response to commands.
    } 
    \label{fig:sparse_and_desired}
\end{figure*}

\subsection{Sparse Delayed Reward Experiments}
Additional experiments were conducted to examine how \UDRL{} is affected as the reward function characteristics change dramatically.
Since \UDRL{} does not use temporal differences for learning, it is reasonable to hypothesize that its behavior may change differently from other algorithms that do.
To test this, we converted environments to their sparse, delayed reward (and thus partially observable) versions by
delaying all rewards until the last step of each episode.
The reward at all other time steps was zero.
A new hyperparameter search was performed for each algorithm on LunarLanderSparse-v2; for other environments we reused the best hyperparameters from the dense setting.
Results for the final 20 runs are plotted in \autoref{fig:sparse_and_desired}.

The baseline algorithms become unstable, very slow or fail completely as expected.
Simple fixes did not work; we failed to train an LSTM DQN on LunarLanderSparse-v2 (while it performed similar to A2C on LunarLander-v2), and A2C with 50 or 100-step returns.
The best performing baseline on this task (A2C with 20-step returns shown in plot) was very sensitive to hyperparameters; very few settings from the search space succeeded in solving the task.
\UDRL{} did not exhibit such sensitivity and its behavior across hyperparameter settings was comparable to the behavior for dense rewards.
It may certainly be possible to solve these tasks by switching to other techniques of standard RL.
Our aim is simply to highlight that \emph{\UDRL{} retained much of its performance in this challenging setting without modification because by design, it directly assigns credit across long time horizons}.
On Inverted Double Pendulum, its performance improved, even approaching the performance of PPO with dense rewards.

\subsection{Different Returns with a Single Agent}
\label{sec:desired}

The \UDRL{} objective simply trains an agent to follow commands compatible with all of its experience, but the learning algorithm in our experiments adds a couple of techniques to focus on higher returns during training in order to make it somewhat comparable to algorithms that focus only on maximizing returns.
This raises the question: do the agents really pay attention to the desired return, or do they simply learn a single policy corresponding to the highest known return?
To answer this, we evaluated agents at the end of training by setting various values of $d^r_0$ and plotting the obtained mean episode return over 100 episodes.
\autoref{fig:sparse_and_desired} (right) shows the result of this experiment on a LunarLander-v2 agent.
It shows a strong correlation ($R\approx0.98$) between obtained and desired returns, even though most of the later stages of training used episodes with returns close to the maximum.
This shows that the agent `remembered' how to act to achieve lower desired returns from earlier training phases.
We note that occasionally this behavior was affected by stochasticity in training, and some agents did not produce very low returns when commanded to do so. 
Additional sensitivity plots for multiple agents and environments are provided in \autoref{sec:sensitivity}.

\section{Related Work}
\label{sec:related}
Episodic Control \citep{lengyel2008hippocampal,blundell2016model} is philosophically similar to our work in taking an alternate perspective on learning to act by directly associating good actions to high returns.
\UDRL{} is conceptually more general, since it attempts to fully model potential relationships between commands and actions, and doesn't target high returns only.
Ideas from Episodic Control were later combined with traditional value-based RL \citep{pritzel2017neural}, which is also a promising future direction for \UDRL{}.

Model-free algorithms that learn value functions \citep{sutton2018} can use experience replay \citep{lin1992} and/or incorporate SL subroutines in many ways \citep{mnih2015,lillicrap2015continuous,mnih2016,schulman2017proximal,oh2018}, but can not take full advantage of SL because they rely on \emph{bootstrapping} which leads to changing targets for the same inputs. 
Both value functions and policies can be goal-conditional \citep{schmidhuber1991b,kaelbling1993,dasilva2012,kupcsik2013,deisenroth2014,schaul2015,pong2018} and learned by relabeling experiences in hindsight \citep{kaelbling1993,andrychowicz2017,rauber2017}, as done in \UDRL{}.
However, what makes it different is that unlike goal-conditioned policies, a behavior function:
(a) can take time-varying desired returns and time horizons as inputs (\autoref{algo:behavior}) as opposed to fixed goals, and 
(b) does not use value functions or policy gradients \citep{williams1992} for training, significantly simplifying the setup for potentially utilizing a variety of SL techniques.

\citet{dosovitskiy2016} showed that Monte Carlo-style future prediction (i.e. without bootstrapping) can be very effective in some high-dimensional environments.
RUDDER \citep{arjona2019rudder} uses gradient-based SL in recurrent networks to perform contribution analysis in order to map rewards to state-action pairs.
{\sc PowerPlay} \citep{schmidhuber2013,srivastava2013c} uses ``task inputs'' to directly produce actions, uses SL on past trajectories to remember previous skills, and does not use value functions. 
However, its task inputs are not constructed systematically based on previous behaviors as we do here.

There are also conceptual connections to ideas formulated around ``RL as inference'', where the solution approach is to infer the values of actions conditioned on the trajectory being optimal \citep{attias2003planning,toussaint2006} (see \citep{levine2018} for a review).
More related is the work of \citet{peters2007} who generalized the expectation maximization framework of \citet{dayan1997} to convert policy search into (on-policy) weighted regression for the \emph{immediate reward} setting.
This was generalized to the episodic case by \citet{wierstra2008a}, and comes closer to offering an SL-like learning setup in each iteration of the algorithm.

Since the first version of this report\footnote{Our work was submitted for blind review to NeurIPS 2019, then accepted to NeurIPS 2019 Deep RL Workshop before becoming public.} we became aware of a few independently developed and related ideas.
We briefly contrast \UDRL{} with these ideas, but recommend a deeper examination to readers.
\citet[defn. 2]{harutyunyan2019} defined an object very similar to our behavior function, but used it differently.
It was always retrained on data from the current policy and used to improve policy gradient estimation.
\citet{lynch2020learning} utilized pure SL to learn a policy conditioned on goals using data generated by a human operator in ``play'' mode.\footnote{This work already appeared online in March 2019 while our experiments were in progress.}
\UDRL{} is more general in that it uses time-varying desires and horizons as inputs instead of fixed goal states, and also uses the same learned behavior function for exploration instead of relying on pre-generated data.
\citet{ghosh2019learning} studied an algorithm that is similar to the one by \citet{lynch2020learning} in using pure SL and conditioning actions on fixed goal states, and similar to ours in that all data for learning is generated by the agent itself.
RCP-R \citep{kumar2019reward} was proposed soon after \UDRL{} and developed based on different motivations, extending recent work on Advantage-Weighted Regression \citep{peng2019advantage}.
At a high level, RCP-R is very similar to \UDRL{} though there are some differences such as use of discount factors instead of explicit horizon conditioning.
Since small implementation details can make a large difference in numerical results on benchmarks, it is difficult to compare results across studies that use different code and evaluation protocols.
Nevertheless, it is notable that \citet{kumar2019reward} motivated their algorithm from the perspective of locally constrained policy improvement and further showed how advantage estimation from traditional RL may be used to improve RCP-R, albeit at the expense of additional complexity.
We discuss a potential inconsistency between training and evaluation in RCP-R in \autoref{subsec:rcpr}.

\section{Discussion}
\label{sec:discussion}

Traditional RL enjoys the benefits of decades of development leading to various algorithms that can, under certain conditions, provably maximize expected returns in arbitrary stochastic environments.
\UDRL{} does not have similar useful theoretical properties yet, so why study a new type of learning for command-following agents that does not build upon well-established principles of stochastic optimal control?
From a practical perspective, many real world problem settings, while not deterministic, are not arbitrarily stochastic either, and learning agents should (and likely do) exploit this.
A broader motivation is that in order to effectively solve the full spectrum of practical problems, as well as to fully understand the behavior of biological learning agents, it is necessary to understand the different ways that agents can encode knowledge and use it to act.
Consider Episodic Control \citep{lengyel2008hippocampal,blundell2016model} that, like \UDRL{}, does not produce optimal expected returns in stochastic environments by itself.
Instead, it models how the hippocampus may support rapid learning in vertebrates by memorizing actions that lead to high returns. 
Algorithms based on different perspectives of learning can avoid problems faced by traditional RL (e.g. due to \emph{the deadly triad}~\citep{sutton2018}), may specialize in specific categories of tasks, and eventually work in concert with other techniques to yield agents with new capabilities.
In the remainder of this section, we briefly discuss some strengths and weaknesses of the ideas presented in this paper.

\paragraph{Avoiding the deadly triad}
Well-known issues arise when off-policy bootstrapping is combined with high-dimensional function approximation.
This combination of techniques --- referred to by \citet{sutton2018} as the \emph{deadly triad} --- can lead to instabilities which are usually addressed by adding a variety of ingredients, though complete remedies remain elusive~\citep{vanhasselt2018}.
\UDRL{}'s central feature is the behavior function, a method to compactly encode knowledge about any set of past behaviors in a new way.
It works fundamentally in concert with high-capacity function approximation to exploit regularities in the environment.
Instead of making predictions about long-term horizons conditioned on a policy (as value functions do), it learns to produce immediate actions conditioned on desired future outcomes.
It opens up the exciting possibility of easily importing a large variety of techniques developed for supervised learning from highly complex data into RL.

\paragraph{Importance of agent architectures}
Our goal throughout this study was to understand the practical viability of the basic principles of \UDRL{} while keeping the experimental setup as simple as possible, particularly in terms of the agent architecture and training strategies such as additional objectives or regularization.
However, after the first phase of experiments, it became clear that it is extremely difficult to reliably train plain neural networks (whose inputs are a concatenation of observations and commands) as useful behavior functions.
An important takeaway from our attempts is that the \UDRL{} objective alone isn't sufficient --- in practice it is crucial that the model class for the behavior function is chosen such that \emph{in any state, the appropriate action to take depends strongly on the command input}.

\paragraph{Avoiding Temporal Distortions}
Many RL algorithms use discount factors that distort true returns.
They are also very sensitive to the frequency of taking actions, limiting their applicability to robot control~\citep{plappert2018}.
\UDRL{} explicitly takes into account observed rewards and time horizons in a precise and natural way, does not assume infinite horizons, and does not suffer from distortions of the RL problem.
In systems that require high-frequency control, even small real-time intervals lead to extremely long horizons, but this is not a conceptual hurdle in \UDRL{} since real-time horizons can directly be used as part of the command inputs.
Our experiments that use step-time horizons as part of commands serve as a successful first step towards such applications.
Note that other algorithms such as evolutionary RL~\citep{moriarty1999} can also avoid these issues in other ways.

\paragraph{Limitations in Stochastic Environments}
\UDRL{} relies on retrospectively generating training data based on achieved commands (instead of what was desired), but it is important to remember this scheme does not yield \emph{optimal} trajectories for the fulfilled commands in stochastic environments.
It only provides example trajectories that \emph{can} fulfill certain commands.
Theoretically analyzing its performance in general stochastic settings, especially  when taking generalization of the behavior function into account, is an open problem.
Note that techniques closer to traditional RL also face theoretical issues when goals are generated in hindsight in stochastic environments (see discussion of HER by \citet[Sec. 2]{plappert2018}).
Like HER, \UDRL{} does not appear to be hindered by them in practice.
A possible explanation is that in many environments with limited stochasticity, the agent can recover from ``unexpected" stochastic transitions by reacting to the actual resulting state and reward observed.

In our experiments, we selected evaluation commands (\autoref{subsec:eval}) assuming that observed returns/horizons (and commands close to them) can be repeated with high probability.
This can work well in many environments with low stochasticity, but in general additional algorithmic ingredients are needed to evaluate the feasibility of commands so that those that are likely to be successful can be generated.

\subsection{Further Research Directions}
\label{sec:future}

There are several directions along which the ideas presented in this paper may be extended.
Using recurrent instead of feedforward neural networks in order to use past observations for selecting actions will be necessary for partially observable settings, and likely useful even in fully observable settings.
New formats of command inputs and architectural modifications tailored to them can substantially improve the inductive biases and learning speed of \UDRL{} agents.
In general, a wide variety of well-known SL techniques for model design, regularization and training can be employed to improve \UDRL{}'s learning stability and efficiency.

Future work should utilize new semantics for command inputs such as ``reach a given goal state in at most $T$ time steps'', 
and strategies for sampling history segments other than just trailing segments.
Similarly, using a constant number of exploratory episodes per iteration decreases sample efficiency and is likely unnecessary.
We also expect that hyperparameters such as the number of optimization updates per iteration can be automatically adjusted during training.

Experiments in this paper used a very simple form of exploration enabled by the behavior function: simply attempt to achieve high returns by generalizing from known states and returns.
This was sufficient to drive learning in the tested environments with a small number of action dimensions and enough stochasticity to help discover new behaviors to learn from.
In other environments, additional forms of undirected (random) and directed exploration should accelerate training and may be necessary.

Finally, there is a vast open space of possible combinations of \UDRL{} and algorithms based on environmental models, temporal-difference learning, optimal control and policy search.
In traditional RL, a successful agent is one that obtains optimal expected returns in stochastic environments.
However, from the \UDRL{} perspective, a successful agent is one that can correctly follow a large variety of commands.
This paper leaves open the question of bridging these two perspectives, but presents evidence suggesting that doing so is likely to be valuable for developing learning agents that are useful in a variety of environments.

\subsection{Broader Impact}
This work falls in the broader category of RL algorithms for autonomously learning to interact with a digital or physical environment.
Potential applications of such algorithms include industrial process control, robotics and recommendation systems.
By attempting to bridge supervised and reinforcement learning, its goal is to make solving RL problems easier and more scalable.
As such, it has the potential to increase both positive and negative impacts of RL research.
An example of potential positive impact is industrial process control to reduce waste and/or energy usage --- industrial combustion is a larger contributor to global greenhouse gas emissions than cars \citep{friedmann2019low}.
Examples of negative impact include unexpected job losses that can have long-term impact on many sections of the society, and feedback loops in online recommendation systems that can maximize engagement with misleading or harmful content.

Our algorithm relies on powerful non-linear function approximators like neural networks in a more direct way than traditional RL algorithms.
Such functions are relatively poorly understood theoretically even in the supervised setting. 
RL agents that use them can behave in unexpected ways, and care must be taken to rigorously test them in safe environments and implement safeguards for preventing and controlling unintended behavior.

\section*{Contributions}
All authors contributed to discussions.
JS proposed the basic principles of \UDRL{} in a companion technical report.
WJ developed the code setup for the baselines and supervised PS.
PS and FM developed early implementations and conducted initial experiments on fully deterministic environments.
RKS developed the final algorithm and architectures, supervised PS and FM, conducted experiments and wrote the paper.

\section*{Acknowledgments}
We thank Paulo E.\ Rauber for discussions during the early stages of this project.
We are grateful to Faustino Gomez for several suggestions that improved this report, and Nihat Engin Toklu for discussions that led to the delayed reward experiments.
Finally we thank Miroslav \v{S}trupl, Saeed Saremi, Kai Arulkumaran, David Ha, Kyunghyun Cho and Sepp Hochreiter for discussions.

\bibliographystyle{humannat}
\bibliography{UD,software}
\newpage
\appendix
\renewcommand*\contentsname{Contents}
\tableofcontents
\newpage

\section{Further Discussion of Related Work}
Almost all RL approaches use or can use supervised learning (SL) subroutines in various ways.
In this section we discuss further related work that may help the reader draw connections between \UDRL{} and other methods.

The discussion in our paper focused on \emph{model-free} algorithms, but certain \emph{model-based} algorithms can also be used to solve RL problems using only SL techniques (i.e.\ without TD-learning or policy gradients).
The general ideas for such algorithms fall under the umbrella of "making the world differentiable" \citep{schmidhuber1990}: learning a differentiable predictive model of the world, that can then be used to directly train a policy by propagating gradients through the world model or other approaches such as planning.
In a certain sense, the behavior function in \UDRL{} and world models are trained in similar fashion on any interactions seen so far without any explicit conditioning on a fixed policy.
The two main differences are that a) the behavior function only encodes algorithmic information about the world that is relevant for following commands, while a world model attempts to encode all information, and b) the behavior function can be directly used to act in the environment without any further planning or training a separate policy.

Certain variants of the RL problem do not consider exploration at all.
Instead, a fixed dataset of interactions with the environment is provided, and the goal is to either mimic the expert policy that generated the data (Imitation Learning \citep{hussein2017}) or to extract the best possible performing policy from the data (Offline/Batch RL \citep{levine2020}).
Many algorithms for these settings are similar to \UDRL{} in that they can use true SL objectives easily, but \UDRL{} goes beyond these settings by exploiting the regularities in the past experience to generate new behavior.

\subsection{Comments on RCP-R}\label{subsec:rcpr}

We briefly discuss an inconsistency between training and \emph{rollout} formulations in RCP-R \citep{kumar2019reward}.
RCP-R labels the rollout trajectory data at each time step ($s_t, a_t$) with the observed value of $Z_t$, defined as the observed reward-to-go from that time step (see step 7, Algorithm 1 and Section 4.1, second para. in their paper).
This labeling is then used for training the current policy.
However, for trajectory rollouts (step 6, Algorithm 1), $Z$ is sampled once at the beginning of the episode and then kept fixed (last line, page 3).
The result is that the joint distribution of states and $Z$ seen during training may be very different from the distribution seen during rollouts for evaluation.
As a simple example, consider an environment where the reward is +1 at each time step.
States that are only encountered later in the trajectories will appear in the training data together with lower values of $Z_t$ compared to states that are always observed earlier.
But during rollouts, the input to the policy will be a fixed high value of $Z$ for all states.

\section{Description of Environments Used}

\begin{figure*}[t]
    \centering
    \begin{subfigure}[b]{0.4\textwidth}
    \includegraphics[width=\textwidth, frame]{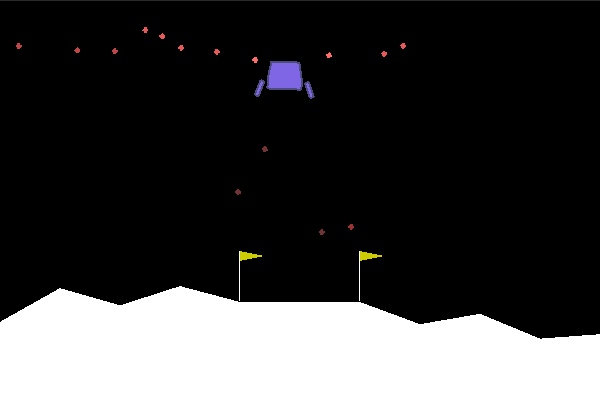}
    \caption{LunarLander-v2}
    \label{fig:renderll}
    \end{subfigure}
    ~
    \begin{subfigure}[b]{0.4\textwidth}
    \includegraphics[width=0.9\textwidth]{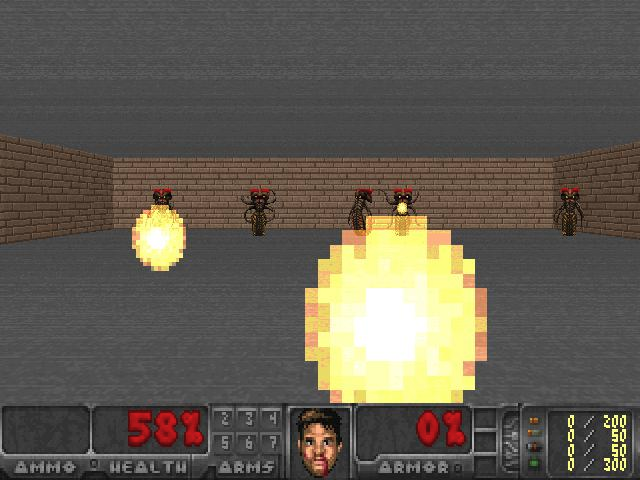}
    \caption{TakeCover-v0}
    \label{fig:rendertc}
    \end{subfigure}
    ~
    \begin{subfigure}[b]{0.4\textwidth}
    \includegraphics[width=\textwidth,height=0.9\textwidth]{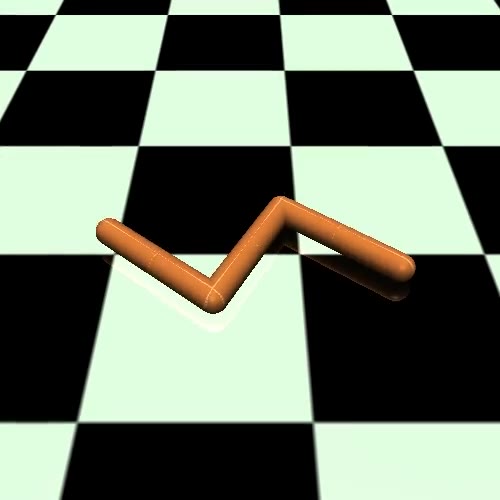}
    \caption{Swimmer-v2}
    \label{fig:rendersw}
    \end{subfigure}
    ~
    \begin{subfigure}[b]{0.4\textwidth}
    \includegraphics[width=0.9\textwidth,height=0.9\textwidth,trim={0 16cm 0 2cm},clip]{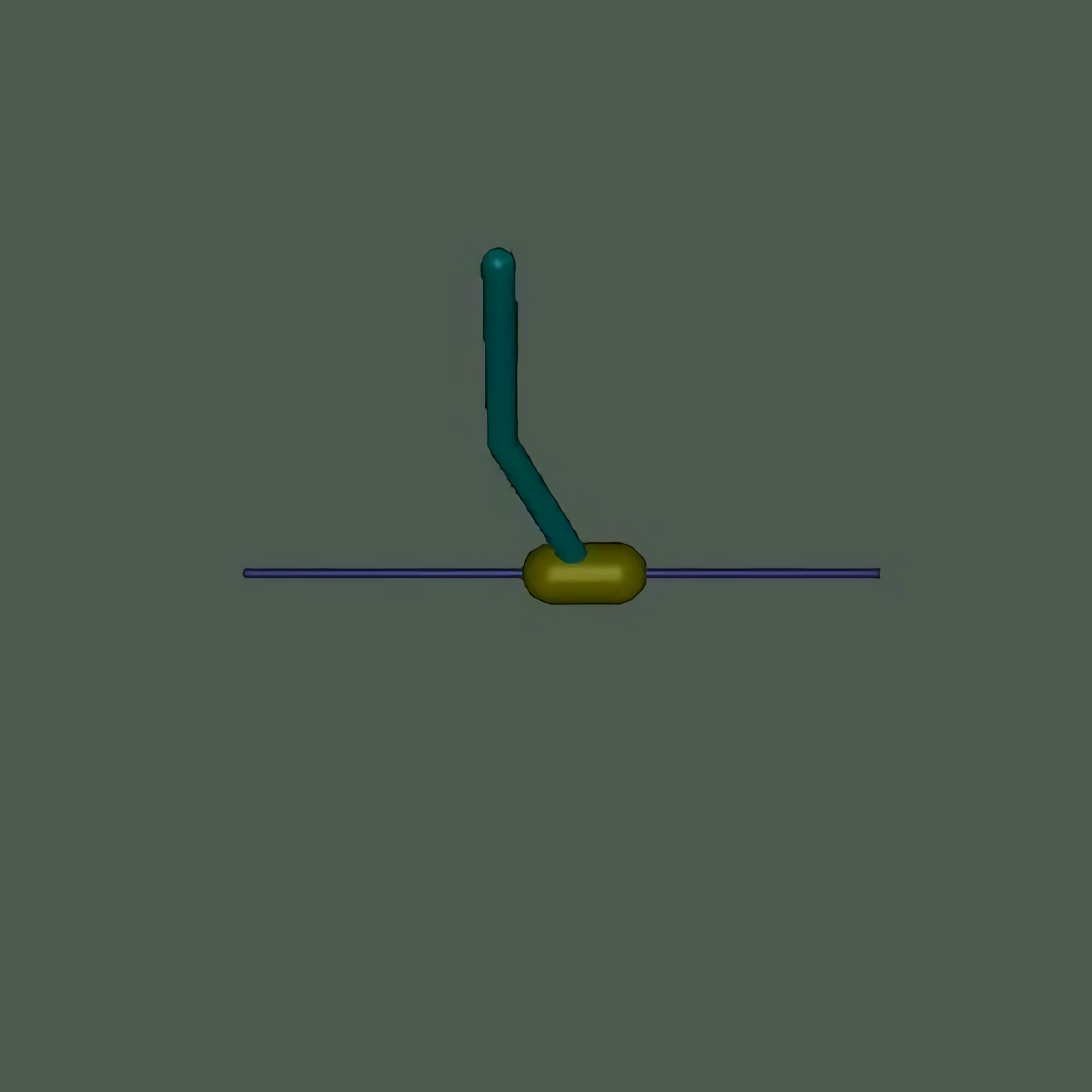}
    \caption{InvertedDoublePendulum-v2}
    \label{fig:renderidp}
    \end{subfigure}
    \caption{Test environments. In TakeCover-v0, the agent observes gray-scale visual inputs down-sampled to 64$\times$64 resolution. In other environments, the observation is a set of state variables.}
    \label{fig:renders}
\end{figure*}

\begin{table}[b]
\centering
\caption{Dimensionality of observations and actions for environments used in experiments.}
\label{tab:envs}
\begin{tabular}[b]{lll}
    \toprule
    Name  & Observations & Actions \\
    \midrule
    LunarLander-v2             &  8           & 4 (Discrete)   \\
    TakeCover-v0               &  8$\times$64$\times$64 & 2 (Discrete)   \\
    Swimmer-v2                 &  8           & 2 (Continuous) \\ 
    InvertedDoublePendulum-v2  &  11          & 1 (Continuous) \\
    \bottomrule
\end{tabular}
\end{table}

\autoref{tab:envs} summarizes the key properties of the environments used in our experiments.
\textbf{LunarLander-v2} (\autoref{fig:renderll}) is a simple Markovian environment available in the Gym RL library \citep{brockman2016} where the objective is to land a spacecraft on a landing pad by controlling its main and side engines.
During the episode the agent receives negative reward at each time step that decreases in magnitude the closer it gets to the optimal landing position in terms of both location and orientation.
The reward at the end of the episode is -100 for crashing and +100 for successful landing.
The agent receives eight-dimensional observations and can take one out of four actions.

\textbf{TakeCover-v0} (\autoref{fig:rendertc}) environment is part of the VizDoom library for visual RL research \citep{kempka2016}. 
The agent is spawned next to the center of a wall in a rectangular room, facing the opposite wall where monsters randomly appear and shoot fireballs at the agent. 
It  must learn to avoid fireballs by moving left or right to survive as long as possible. 
The reward is +1 for every time step that the agent survives, 
so for \UDRL{} agents we always set the desired horizon to be the same as the desired reward, and convert any fractional values to integers.
Each episode has a time limit of 2100 steps, so the maximum possible return is 2100.
Due to the difficulty of the environment (the number of monsters increases with time) and stochasticity, the task is considered solved if the average return over 100 episodes exceeds 750.

Technically, the agent has a non-Markovian interface to the environment,
since it cannot see the entire opposite wall at all times.
To reduce the degree of partial observability, 
the eight most recent visual frames are stacked together to produce the agent observations.
The frames are also converted to gray-scale and down-sampled from an original resolution of 160$\times$120 to 64$\times$64.

\textbf{Swimmer-v2} and \textbf{InvertedDoublePendulum-v2} are environments available in Gym based on the Mujoco engine \citep{todorov2012mujoco}.
In Swimmer-v2, the task is to learn a controller for a three-link robot immersed in a viscous fluid in order to make it swim as far as possible in a limited time budget of 1000 steps.
The agent receives positive rewards for moving forward, and negative rewards proportional to the squared L2 norm of the actions.
The task is considered solved at a return of 360.
In InvertedDoublePendulum-v2, the task is to balance an inverted two-link pendulum by applying forces on a cart that carries it.
The reward is +10 for each time step that the pendulum does not fall, with a penalty of negative rewards proportional to deviation in position and velocity from zero (see source code of Gym for more details).
The time limit for each episode is 1000 steps and the return threshold for solving the task is 9100.

\section{Network Architectures}\label{sec:app-netarchs}

\UDRL{} agents strongly benefit from the use of fast weights controllers \citep{schmidhuber1992learning} --- where outputs of some units are weights (or weight changes) of other connections.
We found that traditional neural architectures where all inputs are simply concatenated \emph{can} yield good performance in some experiments, but they often led to high variation in results across a larger number of random seeds, consequently requiring more extensive hyperparameter tuning.
Therefore, we claim that fast weight architectures are more reliable for \UDRL{} under a limited tuning budget.

For background, \citet{Malsburg:81} and \citet{feldman1982dynamic} were the earliest to highlight the importance of neural networks with rapidly changing weights (as opposed to weights that are fixed after learning).
\citet{hinton1987using} proposed a method where each connection weight was a sum of a slow weight and a fast weight differing in their learning rates and weight decay coefficients, but this has little to do with context-dependent processing where weights or weight changes are outputs of other trainable units, which originated in the work of \citet{schmidhuber1992learning}.

Intuitively, the use of fast weight architectures in \UDRL{} provides a stronger bias towards contextual processing and decision making.
In a traditional network design where observations are concatenated together and then transformed non-linearly, the network can easily learn to ignore command inputs (assign them very low weights) and still achieve lower values of the loss, especially early in training when the experience is less diverse.

Fast weights make it harder to ignore command inputs during training, and even simple variants enable multiplicative interactions between representations.
Such interactions are more natural for representing behavior functions where for the same observations, the agent's behavior should be different depending on the command inputs.
This is supported by past use of such architectures in RL.
For example, \citet{gomez2005evolving} trained non-linear recurrent networks to generate weights for \emph{linear} controllers that could switch between multiple modes of control.
\citet{jayakumar2020multiplicative} trained a single agent to learn better policies for multiple tasks without access to task identities by multiplicatively conditioning processing on the agents internal state.
Fast weights have also been employed in a variety of SL contexts \citep[e.g.][]{ha2016hypernetworks,jia2016dynamic,bertinetto2016learning,schlag2018learning}.

We found a variety of fast weight architectures to be effective during exploratory experiments.
For extensive experiments, we considered two of the simplest options: \emph{gated} and \emph{bilinear} described below.
Considering observation $o \in \Rab{o}{1}$, command $c \in \Rab{c}{1}$, and computed contextual representation $y \in \Rab{y}{1}$, the fast-weight transformations are:

\begin{description}
    \item[Gated]
        \begin{align*}
            g &= \sigma(Uc+p),\\
            x &= f(Vo+q),\\
            y &= x \boldsymbol{\cdot} g.
        \end{align*}
    Here $f$ is a non-linear activation function, $\sigma$ is the sigmoid nonlinearity ($\sigma(x)=(1+e^{-x})^{-1}$), $U \in \Rab{y}{c}$ and $V \in \Rab{y}{o}$ are weight matrices, $p \in \Rab{y}{1}$ and $q \in \Rab{y}{1}$ are biases.
    \item[Bilinear]
        \begin{align*}
            W' &= Uc+p,\\
            b &= Vc+q,\\
            y &= f(Wo+b)
        \end{align*}
    Here $U \in \Rab{(y * o)}{c}$, $V \in \Rab{y}{c}$, $p \in \Rab{(y * o)}{1}$, $q \in \Rab{y}{1}$ and $W$ is obtained by reshaping $W'$ from $(y * o) \times 1$ to $y \times o$.
    Effectively, a linear transform is applied to $o$ where the transformation parameters are themselves produced through linear transformations of $c$.
    This is same as the implementation used by \citet{jayakumar2020multiplicative}.
\end{description}

\citet{jayakumar2020multiplicative} use multiplicative interactions in the last layer of their networks; we used them in the first layer instead (other layers are fully connected) and thus employed an activation function $f$ (typically $y=\max(x, 0)$).
The exceptions are experiments where a convolutional network is used (on TakeCover-v0), where we used a bilinear transformation in the last layer only and did not tune the gated variant.

\subsection{Continuous Control Parameterization}
For environments with continuous action spaces, we chose to parameterize the behavior function to produce an independent normal distribution for each action.
Therefore, the neural network produced a vector of means and a vector of logarithms of the standard deviations as outputs.
To bias the network towards respecting the environment's bounded action space (between -1 and 1), and to avoid numerical issues due to modeling in log space, a few additional modeling choices were made:
the means produced by the network were squashed using the $\tanh$ function, and the log standard deviations were squashed using the sigmoid function, then scaled to lie in the range [-6, 2].

\section{\udRL{} Hyperparameters}
\label{sec:hypers}

\autoref{tab:udrlhypers} summarizes all the hyperparameters for \UDRL{}.

\begin{table*}[ht]
\caption{A summary of UDRL hyperparameters}
\label{tab:udrlhypers}
\centering
\begin{tabularx}{\textwidth}{lX}
\toprule
Name                                & Description \\ 
\midrule
\texttt{batch\_size}                & Number of (input, target) pairs per batch used for training the behavior function \\
\texttt{fast\_net\_option}          & Type of fast weight architecture (\emph{gated} or \emph{bilinear}) \\
\texttt{horizon\_scale}             & Scaling factor for desired horizon input \\
\texttt{last\_few}                  & Number of episodes from the end of the replay buffer used for sampling exploratory commands \\
\texttt{learning\_rate}             & Learning rate for the ADAM optimizer \\
\texttt{n\_episodes\_per\_iter}     & Number of exploratory episodes generated per step of UDRL training \\
\texttt{n\_updates\_per\_iter}      & Number of gradient-based updates of the behavior function per step of UDRL training \\
\texttt{n\_warm\_up\_episodes}      & Number of warm up episodes at the beginning of training \\
\texttt{replay\_size}               & Maximum size of the replay buffer (in episodes)            \\
\texttt{return\_scale}              & Scaling factor for desired horizon input            \\
\bottomrule
\end{tabularx}
\end{table*}

\section{Benchmarking Setup and Hyperparameter Tuning}
\label{sec:tuning}

Random seeds for resetting the environments were sampled from [1\,M, 10\,M) for training, [0.5\,M, 1\,M) for evaluation during hyperparameter tuning,  and [1, 0.5\,M) for final evaluation with the best hyperparameters.
For each environment, random sampling was first used to find good hyperparameters (including network architectures sampled from a fixed set) for each algorithm based on final performance.
With this configuration, final experiments were executed with 20 seeds (from 1 to 20) for each environment and algorithm.
In line with recent work \citep{henderson2018deep,colas2018many}, we found that comparisons based on fewer final seeds were often inaccurate or misleading.

Hyperparameters for all algorithms were tuned by randomly sampling configurations from a pre-defined grid of values, and evaluating each configuration with 2 or 3 different seeds.
The small number of seeds per configuration (to reduce computational cost) is a limitation of this setup since it can lead to performance over-/underestimation.
Agents were evaluated at intervals of 50\,K steps of interaction, and the best hyperparameter configuration was selected based on the mean of evaluation scores for the last 20 evaluations during each experiment, yielding the configurations with the best average performance towards the end of training.

256 configurations were sampled for LunarLander-v2 and LunarLanderSparse-v2, 72 for TakeCover-v0, and 144 for Swimmer-v2 and InvertedDoublePendulum-v2.
Each random configuration of hyperparameters was evaluated with 3 random seeds for LunarLander-v2 and LunarLanderSparse-v2, and 2 seeds for other environments.

Certain hyperparameters that can have an impact on performance and stability in RL were not tuned.
For example, the hyperparameters of the Adam optimizer (except the learning rate) were kept fixed at their default values.
All biases for \UDRL{} networks were zero at initialization, and all weights were initialized using orthogonal initialization \citep{saxe2013exact}. 
No form of regularization (including weight decay) was used for \UDRL{} agents; in principle we expect regularization to improve performance.

Finally, we note that the number of hyperparameter samples evaluated is very small compared to the total grid size.
Thus our experimental setup is a proxy for moderate hyperparameter tuning effort in order to support reasonably fair comparisons, but it is likely that it does not discover the maximum performance possible for each algorithm.

\subsection{Grids for Random Hyperparameter Search}

In the following subsections, we define the lists of possible values for each of the hyperparameters that were tuned for each environment and algorithm.
For traditional algorithms such as DQN etc., any other hyperparameters were left at their default values in the Stable-Baselines library.
The DQN implementation used "Double" Q-learning by default, but additional tricks for DQN that were not present in the original papers were disabled, such as prioritized experience replay.
The best hyperparameter settings obtained are provided as part of the accompanying code in JSON format.

\subsubsection{LunarLander-v2 \& LunarLanderSparse-v2}

\paragraph{Network Architecture}
\begin{itemize}
    \item Network architecture (indicating number of units per layer):
    \newline[32], [32, 32], [32, 64], [32, 64, 64], [32, 64, 64, 64], [64], [64, 64], [64, 128], [64, 128, 128], [64,\,128,\,128,\,128]
\end{itemize}

\paragraph{DQN Hyperparameters}
\begin{itemize}[noitemsep]
    \item Activation function: $[\tanh, \operatorname{relu}]$
    \item Batch Size: [16, 32, 64, 128]
    \item Buffer Size: [10\,000, 50\,000, 100\,000, 500\,000, 1\,000\,000]
    \item Discount factor: [0.98, 0.99, 0.995, 0.999]
    \item Exploration Fraction: [0.1, 0.2, 0.4]
    \item Exploration Final Eps: [0.0, 0.01, 0.05, 0.1]
    \item Learning rate: \texttt{numpy.logspace}(-4, -2, num=101)
    \item Training Frequency: [1, 2, 4]
    \item Target network update frequency: [100, 500, 1000]
\end{itemize}

\paragraph{A2C Hyperparameters}
\begin{itemize}[noitemsep]
    \item Activation function: $[\tanh, \operatorname{relu}]$
    \item Discount factor: [0.98, 0.99, 0.995, 0.999]
    \item Entropy coefficient: [0, 0.01, 0.02, 0.05, 0.1]
    \item Learning rate: \texttt{numpy.logspace}(-4, -2, num=101)
    \item Value function loss coefficient: [0.1, 0.2, 0.5, 1.0]
    \item Decay parameter for RMSProp: [0.98, 0.99, 0.995]
    \item Number of steps per update: [1, 2, 5, 10, 20]
\end{itemize}

\paragraph{\UDRL{} Hyperparameters}
\begin{itemize}[noitemsep]
    \item \texttt{batch\_size}: [512, 768, 1024, 1536, 2048]
    \item \texttt{fast\_net\_option}: [`bilinear', `gated']
    \item \texttt{horizon\_scale}: [0.01, 0.015, 0.02, 0.025, 0.03]
    \item \texttt{last\_few}: [25, 50, 75, 100]
    \item \texttt{learning\_rate}: \texttt{numpy.logspace}(-4, -2, num=101)
    \item \texttt{n\_episodes\_per\_iter}: [10, 20, 30, 40]
    \item \texttt{n\_updates\_per\_iter}: [100, 150, 200, 250, 300]
    \item \texttt{n\_warm\_up\_episodes}: [10, 30, 50]
    \item \texttt{replay\_size}: [300, 400, 500, 600, 700]
    \item \texttt{return\_scale}: [0.01, 0.015, 0.02, 0.025, 0.03]
\end{itemize}

\subsubsection{TakeCover-v0}

\paragraph{Network Architecture}

All networks had four convolutional layers, each with 3$\times$3 filters, 1 pixel input padding in all directions and stride of 2 pixels.
The architecture of convolutional layers (indicating number of convolutional channels per layer) was sampled from [[32, 48, 96, 128], [32, 64, 128, 256], [48, 96, 192, 384]].
The architecture of fully connected layers following the convolutional layers was sampled from [[64, 128], [64, 128, 128], [128, 256], [128, 256, 256], [128, 128], [256, 256]].
Hyperpameter choices for DQN and A2C were the same as those for LunarLander-v2.
For \UDRL{} the following choices were different:

\begin{itemize}[noitemsep]
    \item \texttt{n\_updates\_per\_iter}: [200, 300, 400, 500]
    \item \texttt{replay\_size}: [200, 300, 400, 500]
    \item \texttt{return\_scale}: [0.1, 0.15, 0.2, 0.25, 0.3]
\end{itemize}

\subsubsection{Swimmer-v2 and InvertedDoublePendulum-v2}

\paragraph{Network Architecture}

\begin{itemize}[noitemsep]
    \item Network architecture (indicating number of units per layer):
    \newline[[128], [128, 256], [256, 256], [64, 128, 128, 128], [128, 256, 256, 256]]
\end{itemize}

\paragraph{TRPO Hyperparameters}

\begin{itemize}[noitemsep]
    \item Activation function: $[\tanh, \operatorname{relu}]$
    \item Discount factor: [0.98, 0.99, 0.995, 0.999]
    \item Time steps per batch: [256, 512, 1024, 2048]
    \item Max KL loss threshold: [0.005, 0.01, 0.02]
    \item Number of CG iterations: [5, 10, 20]
    \item GAE factor: [0.95, 0.98, 0.99]
    \item Entropy Coefficient: [0.0, 0.1, 0.2]
    \item Value function training iterations: [1, 3, 5]   
\end{itemize}

\paragraph{PPO Hyperparameters}

\begin{itemize}[noitemsep]
    \item Activation function: $[\tanh, \operatorname{relu}]$
    \item Discount factor: [0.98, 0.99, 0.995, 0.999]
    \item Learning rate: \texttt{numpy.logspace}(-4, -2, num=101)
    \item Number of environment steps per update: [64, 128, 256]
    \item Entropy coefficient for loss: [0.005, 0.01, 0.02]
    \item Value function coefficient for loss: [1.0, 0.5, 0.1]
    \item GAE factor: [0.9, 0.95, 0.99]
    \item Number of minibatches per update: [2, 4, 8]
    \item Number of optimization epochs: [2, 4, 8]
    \item PPO Clipping parameter: [0.1, 0.2, 0.4] 
\end{itemize}

\paragraph{DDPG Hyperparameters}

\begin{itemize}[noitemsep]
    \item Activation function: $[\tanh, \operatorname{relu}]$
    \item Discount factor: [0.98, 0.99, 0.995, 0.999]
    \item Sigma for OU noise: [0.1, 0.5, 1.0]
    \item Observation normalization: [False, True]
    \item Soft update coefficient: [0.001, 0.002, 0.005]
    \item Batch Size: [128, 256]
    \item Return normalization: [False, True]
    \item Actor learning rate: \texttt{numpy.logspace}(-4, -2, num=101)
    \item Critic learning rate: \texttt{numpy.logspace}(-4, -2, num=101)
    \item Reward scale: [0.1, 1, 10]
    \item Buffer size: [50\,000, 100\,000]
    \item Probability of random exploration: [0.0, 0.1] 
\end{itemize}

\paragraph{\UDRL{} Hyperparameters}

\begin{itemize}[noitemsep]
    \item \texttt{batch\_size}: [256, 512, 1024]
    \item \texttt{fast\_net\_option}: [`bilinear', `gated']
    \item \texttt{horizon\_scale}: [0.01, 0.02, 0.03, 0.05, 0.08]
    \item \texttt{last\_few}: [1, 5, 10, 20]
    \item \texttt{learning\_rate}: [0.0001, 0.0002, 0.0004, 0.0006, 0.0008, 0.001]
    \item \texttt{n\_episodes\_per\_iter}: [5, 10, 20]
    \item \texttt{n\_updates\_per\_iter}: [250, 500, 750, 1000]
    \item \texttt{n\_warm\_up\_episodes}: [10, 30, 50]
    \item \texttt{replay\_size}: [50, 100, 200, 300, 500]
    \item \texttt{return\_scale}: [0.01, 0.02, 0.05, 0.1, 0.2]
\end{itemize}

\subsection{Tuned Hyperparameter Values}

\autoref{tab:tuned-hypers} reports the tuned values of \UDRL{} hyperparameters for each of the tasks.
We note that small deviations in the code, library versions or hardware accelerators can lead to some variations in the obtained results.

\begin{table}[h]
\centering
\footnotesize
\caption{Tuned hyperparameter values for \UDRL{} tuned using random search.}
\label{tab:tuned-hypers}
\begin{tabular}{@{}llllll@{}}
\toprule
                                & LunarLander-v2        & Swimmer-v2 & \begin{tabular}[c]{@{}l@{}}Inverted\\Double\\Pendulum-v2\end{tabular} & TakeCover-v2                                                                      & \begin{tabular}[c]{@{}l@{}}LunarLander\\Sparse-v2\end{tabular} \\
\midrule
Network architecture            & [64, 128, 128]        & [128]      & 128                       & \begin{tabular}[c]{@{}l@{}}Conv: [48, 96, 192, 384]\\ FC:\quad\,\,[256, 256]\end{tabular} & [64, 128, 128, 128]  \\
\texttt{batch\_size}            & 768                   & 512        & 256                       & 768                                                                               & 512                  \\
\texttt{fast\_net\_option}      & bilinear              & gated      & gated                     & bilinear                                                                          & gated                \\
\texttt{horizon\_scale}         & 0.03                  & 0.01       & 0.01                      & 0.1                                                                               & 0.015                \\
\texttt{last\_few}              & 100                   & 1          & 5                         & 75                                                                                & 25                   \\
\texttt{learning\_rate}         & 0.0008709635899560805 & 0.0002     & 0.0006                    & 0.0007244359600749898                                                             & 0.000501187          \\
\texttt{n\_episodes\_per\_iter} & 20                    & 20         & 20                        & 20                                                                                & 20                   \\
\texttt{n\_updates\_per\_iter}  & 150                   & 750        & 500                       & 500                                                                               & 250                  \\
\texttt{n\_warm\_up\_episodes}  & 50                    & 30         & 50                        & 50                                                                                & 30                   \\
\texttt{replay\_size}           & 600                   & 50         & 300                       & 300                                                                               & 500                  \\
\texttt{return\_scale}          & 0.015                 & 0.01       & 0.01                      & 0.1                                                                               & 0.015                \\
Std.\ dev.\ for warmup  &                       & 0.3        & 0.35                      &                                                                                   &  \\
\bottomrule
\end{tabular}
\end{table}

\section{Additional Plots}
\subsection{SwimmerSparse-v2 results}

\begin{figure}
\centering
\includegraphics[width=0.6\textwidth]{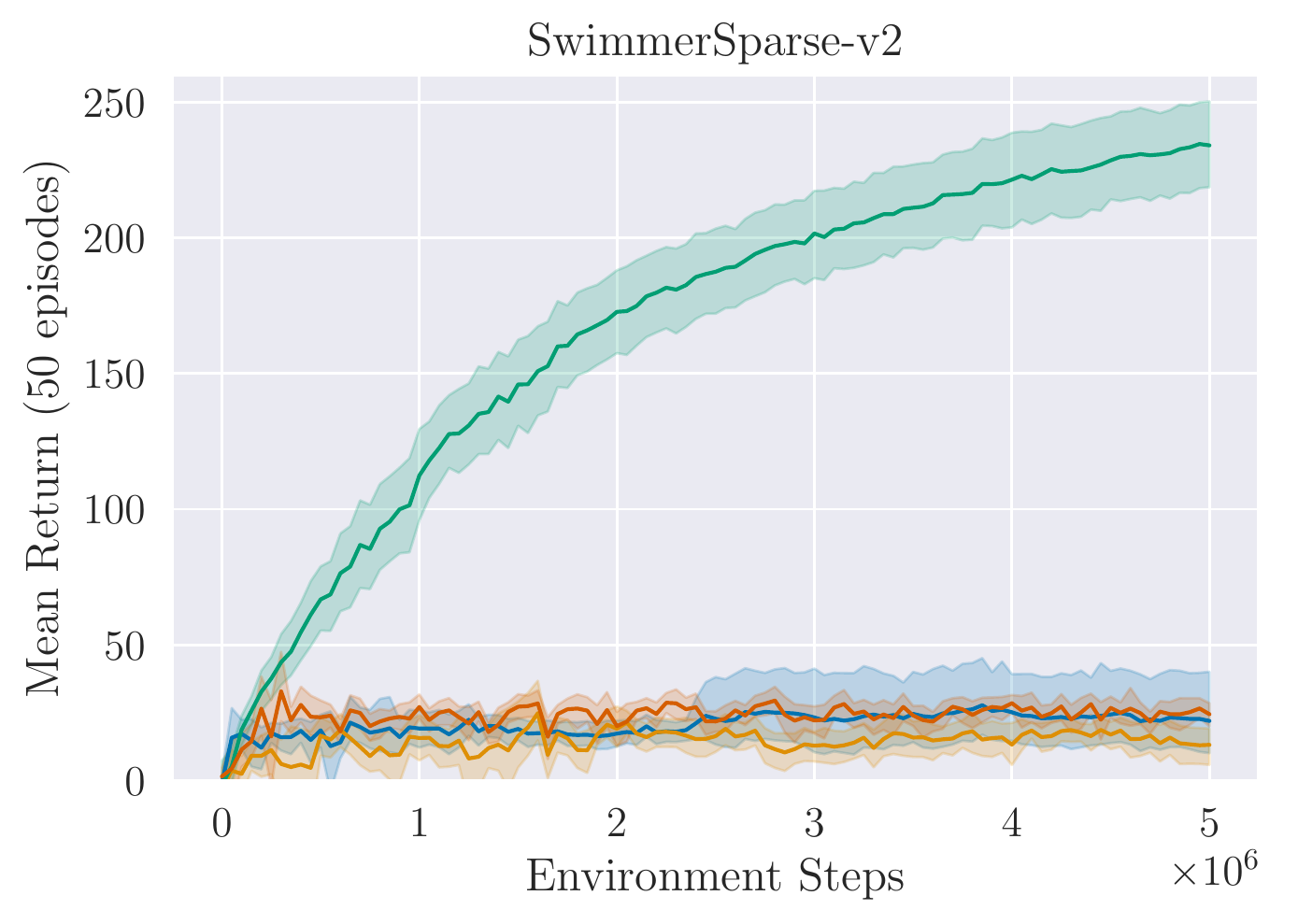}
\caption{Results on SwimmerSparse-v2. Blue: TRPO, Yellow: PPO, Red: DDPG, Green: \UDRL{}.}
\label{fig:swsparse}
\end{figure}

\autoref{fig:swsparse} presents the results on SwimmerSparse-v2, the sparse delayed reward version of the Swimmer-v2 environment.
Similar to other environments, the key observation is that \UDRL{} retained much of its performance \textbf{without modification}.
The hyperparameters used were same as the dense reward environment.

\subsection{Sensitivity to Initial Commands}\label{sec:sensitivity}

\begin{figure*}[t]
    \centering
    \begin{subfigure}[b]{0.49\textwidth}
    \includegraphics[width=\textwidth]{figures/llv2_obtained_vs_desired_return_1.pdf}
    \caption{}
    \label{subfig:desired-llv2}
    \end{subfigure}
    ~
    \begin{subfigure}[b]{0.49\textwidth}
    \includegraphics[width=\textwidth]{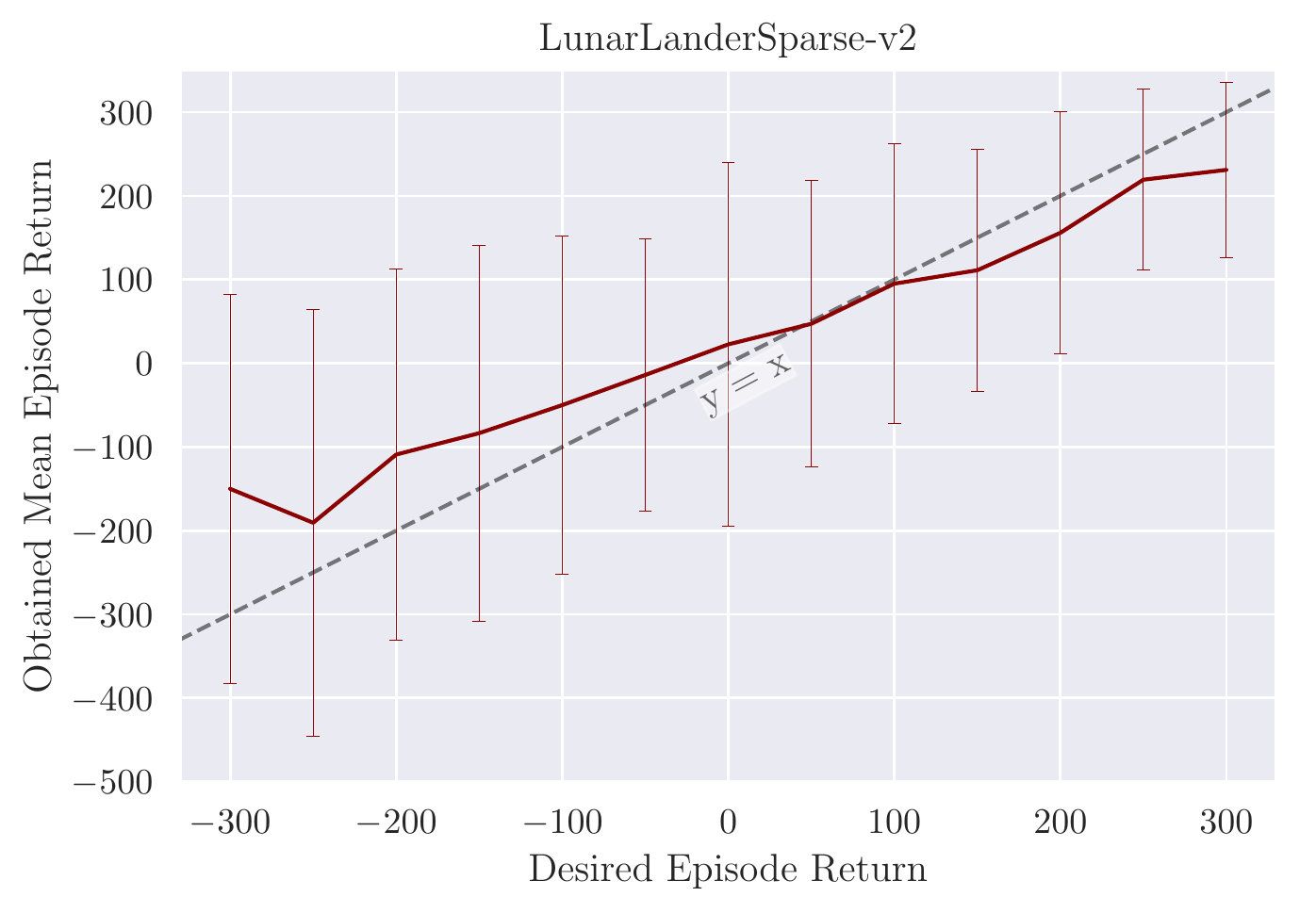}
    \caption{}
    \label{subfig:desired-lls9}
    \end{subfigure}
    \\
    \begin{subfigure}[b]{0.49\textwidth}
    \includegraphics[width=\textwidth]{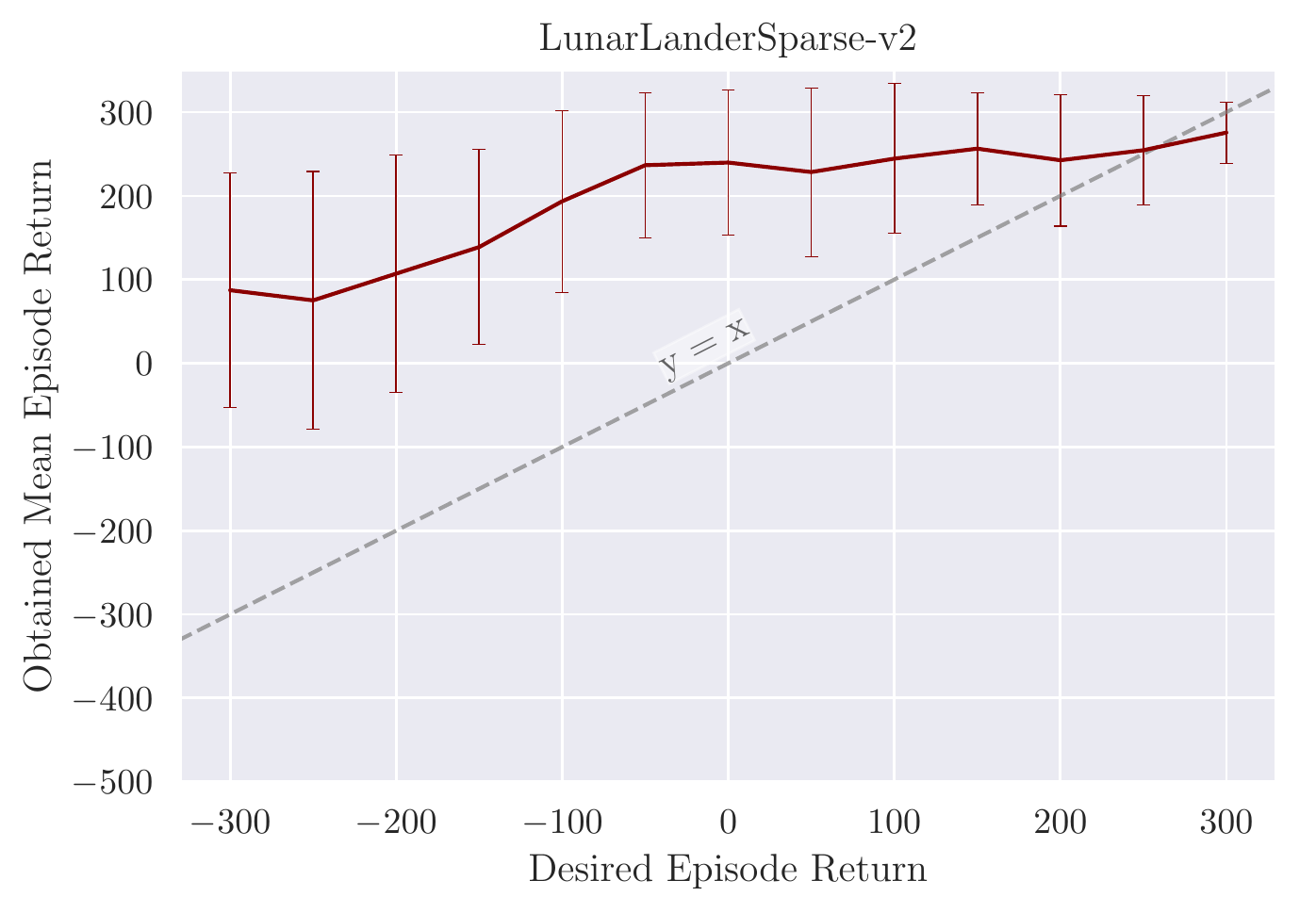}
    \caption{}
    \label{subfig:desired-lls1}
    \end{subfigure}
    ~
    \begin{subfigure}[b]{0.49\textwidth}
    \includegraphics[width=\textwidth]{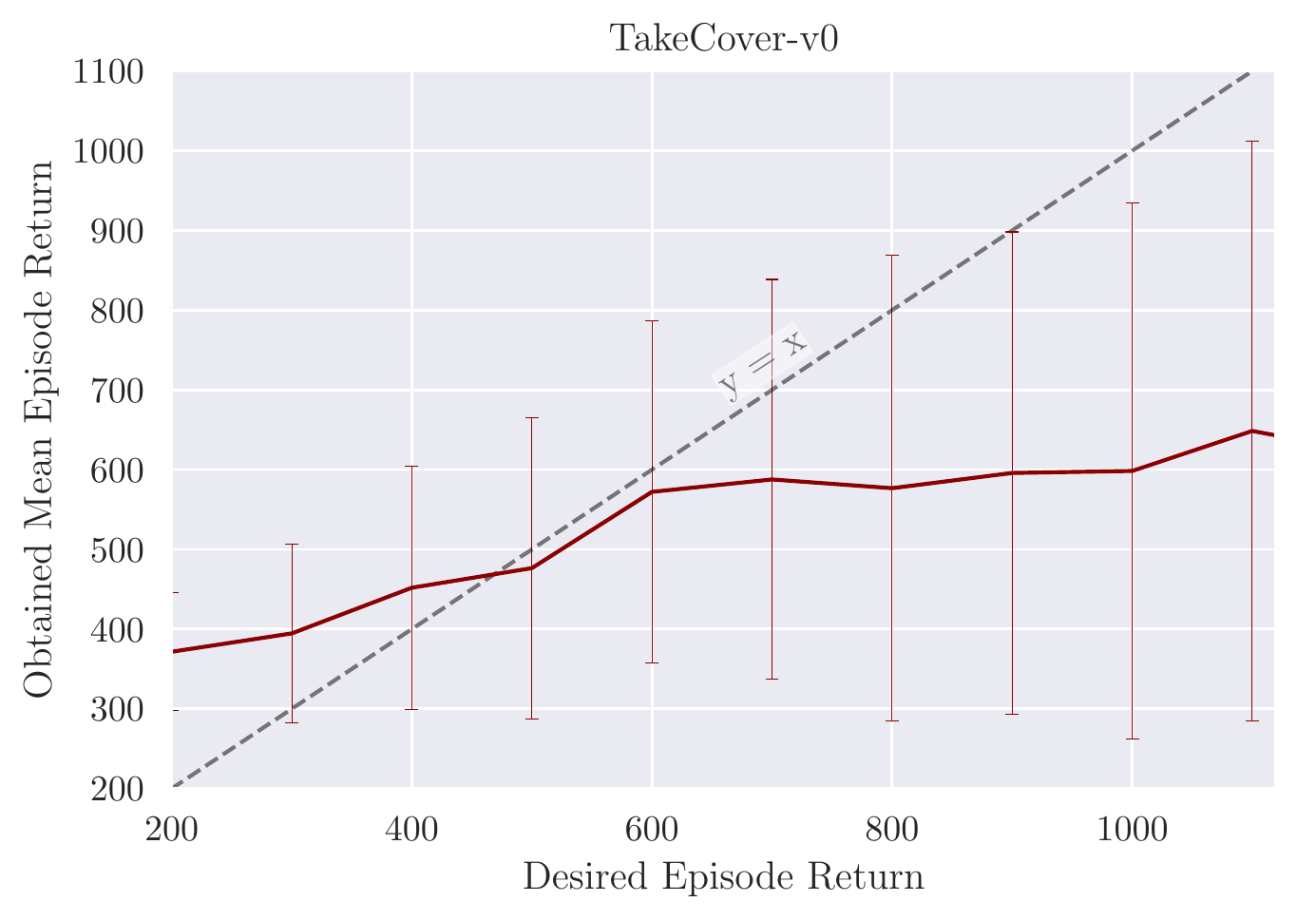}
    \caption{}
    \label{subfig:desired-tc}
    \end{subfigure}
    \\
    \begin{subfigure}[b]{0.49\textwidth}
    \includegraphics[width=\textwidth]{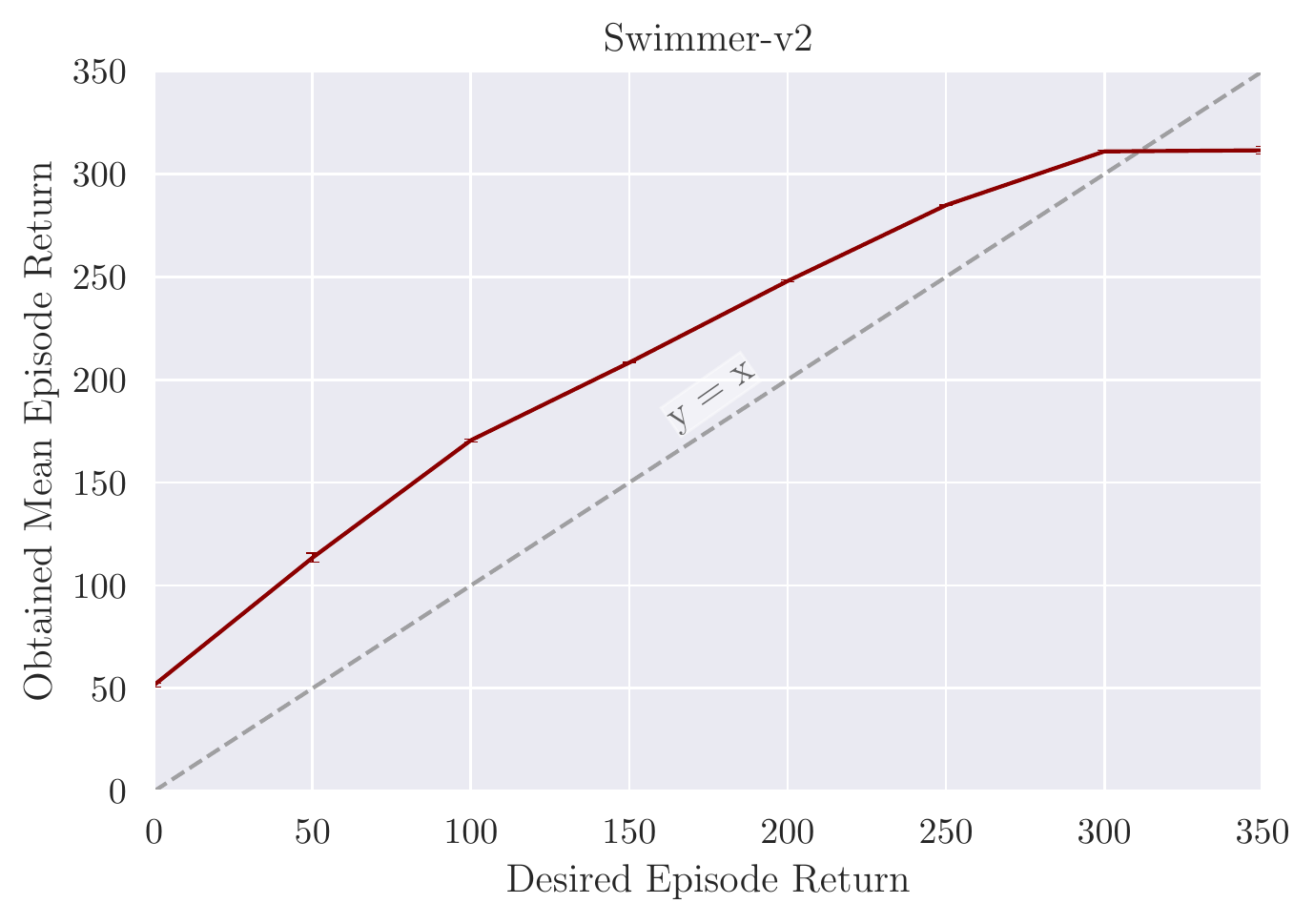}
    \caption{}
    \label{subfig:desired-sw}
    \end{subfigure}
    ~
    \begin{subfigure}[b]{0.49\textwidth}
    \includegraphics[width=\textwidth]{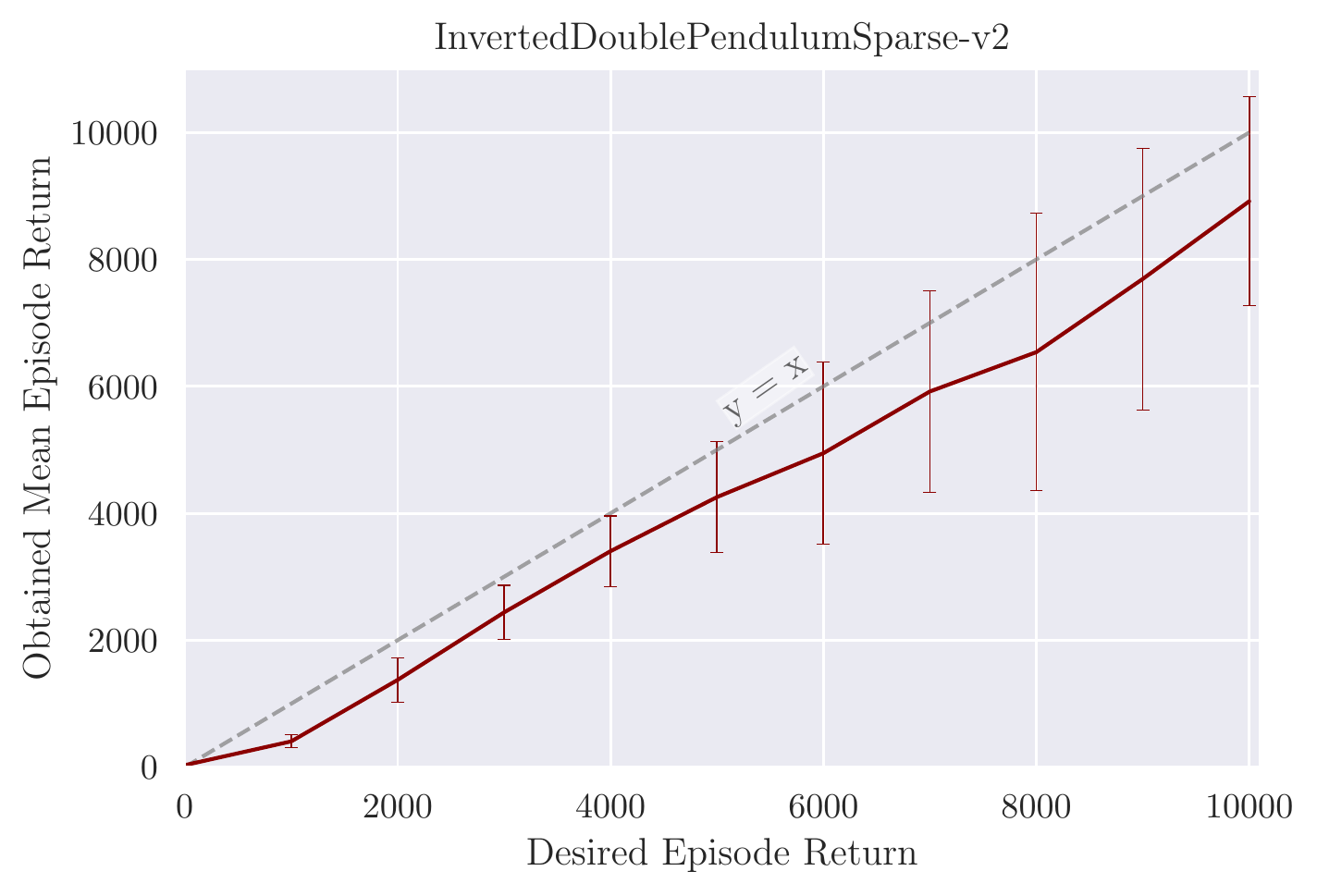}
    \caption{}
    \label{subfig:desired-idp}
    \end{subfigure}
    
    \caption{Obtained vs. desired episode returns for UDRL agents at the end of training. Each evaluation consists of 100 episodes. Error bars indicate standard deviation from the mean. Note the contrast between (b) and (c): both are agents trained on LunarLanderSparse-v2. The two agents differ only in the random seed used for the training procedure, showing that variability in training can lead to different sensitivities at the end of training.}
    \label{fig:desired}
\end{figure*}

This section includes additional evaluations of the sensitivity of \UDRL{} agents at the end of training to a series of initial commands (see Section 3.4 in the main paper).
\Cref{subfig:desired-llv2,subfig:desired-lls9,subfig:desired-sw,subfig:desired-idp} show a strong correlation between obtained and desired returns for randomly selected agents on LunarLander-v2 and LunarLanderSparse-v2.
Notably, \autoref{subfig:desired-lls1} shows another agent trained on LunarLanderSparse-v2 that obtains a return higher than 200 for most values of desired returns, and only achieves lower returns when the desired return is very low. 
This indicates that stochasticity during training can affect how trained agents generalize to different commands, and suggests another direction for future investigation.

A possible complication in this evaluation is that it is unclear how to set the value of the initial desired horizon $d^h_0$ for various values of $d^r_0$.
This is easier in some environments: in TakeCover-v0, we set $d^h_0$ equal to $d^r_0$.
Similarly, for InvertedDoublePendulumSparse-v2 where the agent gets a reward of +10 per step, we set $d^h_0$ = $d^r_0 / 10$. 
This does not take the position and velocity penalties into account, but is sufficiently reasonable.
For other environments, we simply use the $d^h_0$ value at the end of training for all desired returns.
In general, the agent's lifelong experience can be used to keep track of realistic values of $d^h_0$ and $d^r_0$, which may be dependent on the initial state.

In the TakeCover-v0 environment, it is rather difficult to achieve precise values of desired returns. 
Stochasticity in the environment (the monsters appear randomly and shoot in random directions) and increasing difficulty over the episode imply that it is not possible to achieve lower returns than 200 and it becomes progressively harder to achieve higher mean returns.
The results in \autoref{subfig:desired-tc} reflect these constraints. 
Instead of increased values of mean returns, we observe higher standard deviation for higher values of desired return.

\section{Software \& Hardware}
Our setup directly relied upon the following open source software:

\begin{itemize}[noitemsep]
    \item Gym 0.15.4 \citep{brockman2016}
    \item Matplotlib \citep{hunter2007}
    \item Numpy 1.18.1 \citep{walt2011,oliphant2015}
    \item OpenCV \citep{bradski2000}
    \item Pytorch 1.4.0 \citep{paszke2017}
    \item Ray Tune 0.6.6 \citep{liaw2018}
    \item Sacred 0.7.4 \citep{greff2017sacred}
    \item Seaborn \citep{waskom2018}
    \item Stable-Baselines 2.9.0 \citep{stable-baselines}
    \item Vizdoom 1.1.6 \citep{kempka2016}
    \item Vizdoomgym \citep{hakenes2018}
\end{itemize}

For all LunarLander and TakeCover experiments, \texttt{gym==0.11.0}, \texttt{stable-baselines==2.5.0}, and \texttt{pytorch==1.1.0} were used.
Mujoco 1.5 was used for continuous control tasks.

All experiments were run on cloud computing resources with Intel Xeon processors, the exception being some TakeCover experiments which were run on local hardware with Nvidia V100 GPUs. When cloud resources were used, Nvidia P100 GPUs were used for TakeCover experiments. Each experiment occupied one or two vCPUs, and 33\% GPU capacity (if used).

\end{document}